\documentclass[runningheads,a4paper]{llncs}
\usepackage{amssymb}
\usepackage{subfigure}
\usepackage{graphicx, wrapfig}
\usepackage[font=scriptsize,skip=0pt]{caption}
\usepackage{comment}
\usepackage{url}
\usepackage{amsmath}
\usepackage{bm}
\usepackage{xcolor}

\newcommand{\real}{\mathbb{R}}

\newcommand{\cG}{\mathcal{G}}

\newcommand{\mcN}{\mathcal{N}}

\newcommand{\mcL}{\mathcal{L}}

\usepackage{xcolor}

\def \xb{\mathbf{x}}

\def \hb{\mathbf{h}}

\def \cb{\mathbf{c}}

\urldef{\mailsa}\path|{somdatta_goswami@brown.edu, aniruddha_bora@brown.edu, yuy214@lehigh.edu,  george_karniadakis@brown.edu}|    
\newcommand{\keywords}[1]{\par\addvspace\baselineskip
\noindent\keywordname\enspace\ignorespaces#1}

\begin{document}

\mainmatter  
\title{Physics-Informed Deep Neural Operator Networks}
\titlerunning{Physics-Informed Deep Neural Operator Networks}
\author{Somdatta Goswami \inst{1} \and Aniruddha Bora \inst{1} \and Yue Yu \inst{2} \and George Em Karniadakis\thanks{Corresponding author}\inst{1,3}}

\authorrunning{Physics-Informed Deep Neural Operator Networks}
\institute{Division of Applied Mathematics, Brown University, Providence, RI, USA. \\\email{\{somdatta\_goswami, aniruddha\_bora, george\_karniadakis\}@brown.edu}\and
Department of Mathematics, Lehigh University, Bethlehem, PA, USA.\\\email{yuy214@lehigh.edu}\and
School of Engineering, Brown University, Providence, RI, USA}

\maketitle

\begin{abstract}
Standard neural networks can approximate general nonlinear operators, represented either explicitly by a combination of mathematical operators, \textit{e.g.}, in an advection-diffusion-reaction partial differential equation, or simply as a black box, \textit{e.g.}, a system-of-systems. The first neural operator was the Deep Operator Network (DeepONet), proposed in 2019 based on rigorous approximation theory. Since then, a few other less general operators have been published, e.g., based on graph neural networks or Fourier transforms. For black box systems, training of neural operators is data-driven only but if the governing equations are known they can be incorporated into the loss function during training to develop {\em physics-informed} neural operators. Neural operators can be used as surrogates in design problems, uncertainty quantification, autonomous systems, and almost in any application requiring real-time inference. Moreover, independently pre-trained DeepONets can be used as components of a complex multi-physics system by coupling them together with relatively light training. Here, we present a review of DeepONet, the Fourier neural operator, and the graph neural operator, as well as appropriate extensions with feature expansions, and highlight their usefulness in diverse applications in computational mechanics, including porous media, fluid mechanics, and solid mechanics. 

\end{abstract}

\section{Introduction}
Physics-informed neural networks (PINNs) have transformed the way we model the behavior of physical systems for which we have available some measurements and at least a parameterized partial differential equation (PDE) to provide additional information in a semi-supervised type of learning \cite{raissi2019physics,karniadakis2021physics,samaniego2020energy}. They can solve ill-posed problems that may lack boundary conditions, \textit{e.g.}, thermal boundary conditions in heat transfer problems \cite{cai2021physics} or discover voids and defects in materials based only on a handful of displacements \cite{zhang2022analyses}, or obtain the failure pattern \cite{goswami2020adaptive}. Despite their effectiveness, PINNs are trained for specific boundary and initial conditions, as well as loading or source terms, and require expensive optimization during inference. Therefore, they are not particularly effective for other operating conditions and real-time inference, although transfer learning can somewhat alleviate this limitation \cite{goswami2020transfer,goswami2022deep}. What we need in engineering disciplines such as design, control, uncertainty quantification, robotics, etc. is a generalized version of PINNs that can infer the system's response in real-time for many different boundary/initial conditions and loadings, without further training or perhaps with very light training. This will lead to speed-up factors of thousands compared to conventional numerical solvers, \textit{e.g.}, CFD or solid mechanics simulators. 

The neural operators, introduced in 2019 in the form of DeepONet \cite{lu2021_deeponet}, fulfill this promise. Training is performed offline in a predefined input space and hence inference is very fast since no further training is required as long as the new conditions are inside the input space. For arbitrary inputs, which are out of the distribution (OOD), further training is required but this may be relatively light if the input space is sampled sufficiently. We note here that a neural operator is very different from a reduced order model (ROM) that is restricted to a very small subset of conditions and lacks generalization properties due to the under-parameterization in such methods \cite{kontolati2022influence,geelen2022operator}. Another important property of DeepONet is that it is based on a universal approximation theorem for operators \cite{chen_chen_approximation,lu2021_deeponet}, and more recent theoretical work \cite{lanthaler2022error} has shown that DeepONet can break the curse of dimensionality in the input space, unlike approaches based on ROM for parameterized PDEs \cite{riffaud2021dgdd}.   

\begin{figure}
  \begin{center}
 \vspace*{-0.8cm}
    \includegraphics[width=\textwidth]{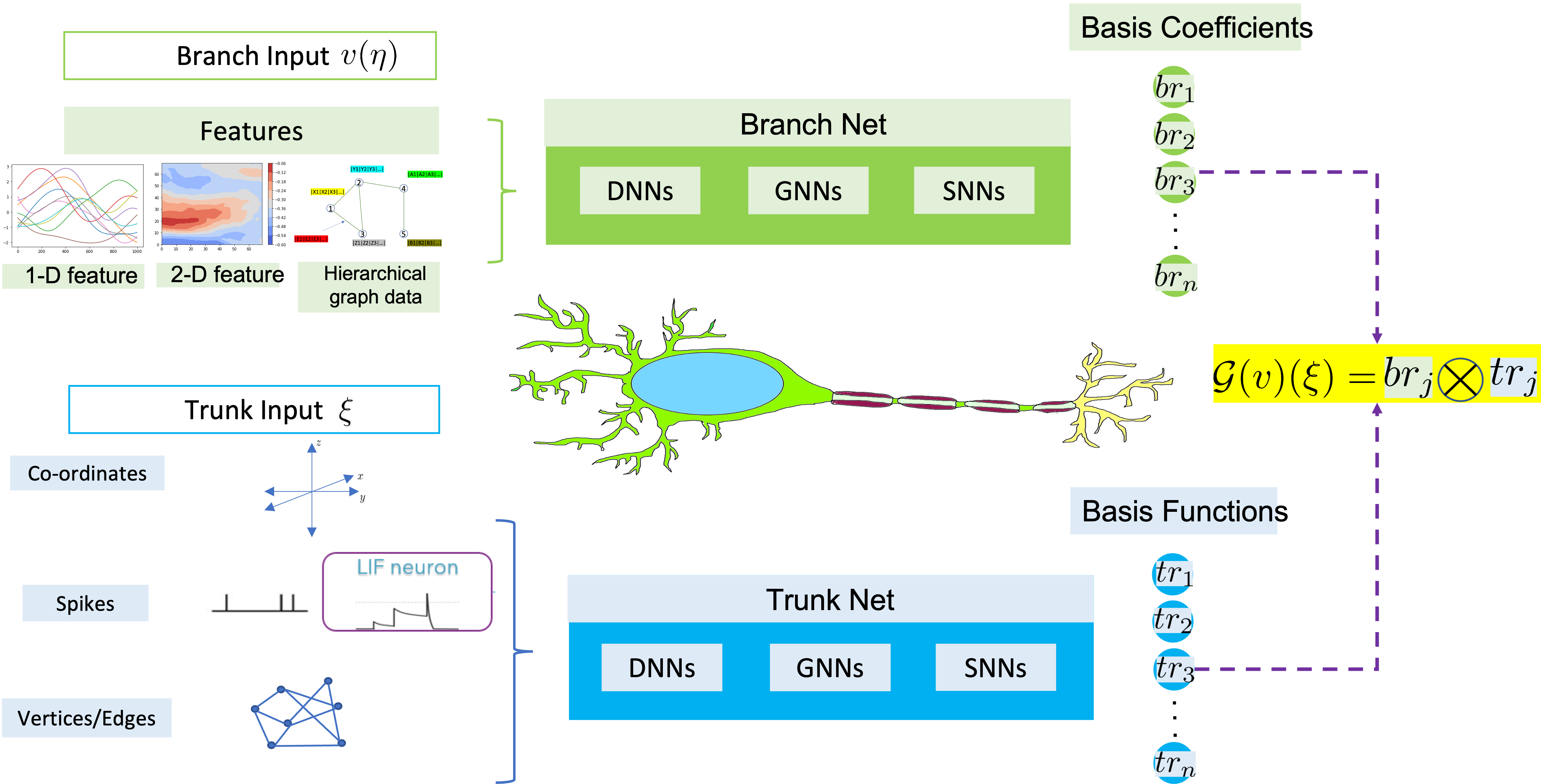}
  \end{center}
  \vspace*{-0.1cm}
  \caption{Schematic representation of the deep operator network (DeepONet) at inception in 2019. It consists of two deep neural networks with flexible architectures, and the dot product of the outputs of the two networks represents the function of the solution. The color-coding indicates corresponding resemblance of computational operations with a human neuron. The schematic shows that different inputs can be employed in the branch and trunk networks, including graphs and spikes.}
  \label{fig:don} 
\end{figure}

Figure \ref{fig:don} represents the schematic of a polymorphic DeepONet and its resemblance to a human neuron. An operator network is made up of two deep neural networks (DNNs): branch and trunk. The branch and trunk networks are analogous to synchronized dendritic branch and axonal spiking. The result is a non-linear operator that can be used to approximate any function defined as an input in the branch network and evaluated at the locations specified in the trunk network. In 2020, work on \emph{graph kernel network} (GKN) for PDE led to another type of operator regression \cite{Li_GKN}. Subsequently, the same group proposed a different architecture in which they formulated the operator regression by parameterizing the integral kernel directly in Fourier space and named it \emph{Fourier Neural Operator} (FNO) \cite{Li_FNO}. All these versions are different realizations of DeepONet if appropriate changes in the trunk and branch are imposed, see Figure 1 and also \cite{lu2022comprehensive}. 

Consider an operator $\mathcal{G}$, that maps from the input function $v$ to the output function $u$, i.e., $\mathcal{G}: v \rightarrow u$. DeepONet tries to learn the operator $\mathcal{G}$ by approximating the basis function for expressing the output functional space. In this chapter, we will discuss in detail the three aforementioned neural operators, their extensions, and put forward numerical examples to illustrate the usage and limitations of each of these approaches. The physics of the problem is enforced using the labelled input-output dataset pairs for the conventional architecture of the proposed operators. In related sections, variants of the operators that use the physics of the problem (and require little to no labeled data) to train the network, are also covered.

This chapter is organized as follows. In Section \ref{sec:deeponet}, we discuss the deep neural operator (DeepONet), its extensions and variants that have been developed and used for different problems setups. In Section \ref{sec:fno}, we describe the Fourier neural operator (FNO) architecture and present the extensions of the operator to deal with complex geometries, problems defined in different input and output spaces, and also the physics-informed version of the operator. In Section \ref{sec:gno}, we introduce the graph neural operator (GNO), its components, and non-local kernel networks. In Section \ref{sec:application}, we compare the performance of the studied models for several examples from the literature. Finally, we summarize our observations and provide concluding remarks in Section \ref{sec:summary}.

\section{DeepONet and its extensions}
\label{sec:deeponet}

DeepONet is based on the universal approximation theorem for operators, which is defined in Theorem \ref{thm:operator-univ-apprx-nn-uniform}. The conventional DeepONet architecture consists of two DNNs: one encodes the input function, $\mathbf v(\eta)$, at fixed sensor points (branch net) to provide the coefficients, and another encodes the information related to the spatio-temporal coordinates of the output function (trunk net) to provide the basis functions. The goal of DeepONet is to learn the operator $\mathcal G(\mathbf{v})$ that can be evaluated at continuous spatio-temporal coordinates $\xi$ (input to the trunk net). The output of the DeepONet for a specified input vector, $\mathbf{v}_i$, is a scalar-valued function of $\xi$ expressed as $\mathcal G_{\mathbf{\theta}}(\mathbf{v}_i)(\xi)$, where $\mathbf{\theta} = \left(\mathbf W, \mathbf b \right)$ includes the trainable parameters (weights $\mathbf W$, and biases $\mathbf b$) of the networks. To work with the input function of the branch net, it must be discretized in a finite-dimensional space using a finite number of points, termed as sensors. We specifically evaluate $\mathbf v_i$ at $m$ fixed sensor locations $\{\eta_1, \eta_2, \dots, \eta_m\}$ to obtain the point-wise evaluations, $\mathbf{v}_i=\{\mathbf{v}_{i}(\eta_1), \mathbf{v}_{i}(\eta_2), \ldots, \mathbf {v}_{i}(\eta_m)\}$, which are used as input to the branch net. The trunk net receives the spatial and temporal coordinates \textit{e.g.}, $\xi = \{x_i, y_i, t_i\}$ and evaluates the solution operator to compute the loss function. The solution operator for an input realization, $\mathbf v_1$, can be expressed as: 
\begin{equation}\label{eq:output_deeponets}
    \begin{split}
      \mathcal G_{\mathbf\theta}(\mathbf{v}_1)(\mathbf \xi) &= \sum_{i = 1}^p br_i \cdot tr_i \\
      &=\sum_{i = 1}^{p}br_i(\mathbf{v}_{1}(\eta_1), \mathbf{v}_{1}(\eta_2), \ldots, \mathbf {v}_{1}(\eta_m))\cdot tr_i(\xi),   
\end{split}
\end{equation}
where ${br_1, br_2, \ldots, br_p}$ are outputs of the branch net, and ${tr_1, tr_2, \ldots, tr_p}$ are outputs of the trunk net. 

\begin{theorem}[\textbf{Generalized universal approximation theorem for operators}  \cite{lu2021_deeponet}]\\
\label{thm:uat-cite-deeponets}
\label{thm:operator-univ-apprx-nn-uniform}
Suppose that $X$ is a Banach space, $K_1 \subset X$, $K_2 \subset \mathbb{R}^{D_{in}}$ are two compact sets in $X$ and $\mathbb{R}^{D_{in}}$, respectively, $V$ is a compact set in $C(K_1)$  
Assume that $\cG:\, V\to C(K_2) $ is a nonlinear continuous operator. 
Then for any $\epsilon>0$, there exist positive integers $m,p$, 
branch nets $br_k$, and trunk nets  $tr_k$, and 
$\eta_1, \eta_2, \cdots, \eta_m \in K_1$, such that
\begin{equation*}
\sup_{v\in V} \sup_{\xi\in K_2}	\left|\cG(v)(\xi) -\sum_{k=1}^p \underbrace{br_k(v(\eta_1),v(\eta_2),\ldots,v(\eta_m))}_{branch}
	\underbrace{tr_k(\xi)}_{trunk}
	\right|<\epsilon .
\end{equation*}
Furthermore, the functions $br_k$ (outputs of the branch network) and $tr_k$ (outputs of the trunk network) can be chosen as diverse classes of neural networks, satisfying the classical universal approximation theorem of functions, e.g., fully-connected neural networks (FNN), residual neural networks, and convolution neural networks (CNN).
\end{theorem}

This theorem was proved in \cite{chen1995universal} with two-layer neural networks. Also, the theorem holds when the Banach space $C(K_1)$ is replaced by $L^q(K_1)$ and $C(K_2)$ replaced by $L^r(K_2)$, $q,r\geq 1$. We notice that in Eq.~\eqref{eq:output_deeponets}, the last layer of each $br_k$ branch network is bias-free. Although bias is not a requirement in Theorem \ref{thm:operator-univ-apprx-nn-uniform}, adding bias may improve performance by lowering the generalization error \cite{lu2021_deeponet}. The trainable parameters of a data-driven DeepONet, represented by $\mathbf{\theta}$ in Eq.~\eqref{eq:output_deeponets}, are obtained by minimizing a loss function, which is expressed as:
\begin{equation}\label{eq:loss_func_deeponet}
    \mathcal L(\mathbf{\theta}) = \frac{1}{N}\sum_{i = 1}^{N} w_i|u_i(\xi) - \mathcal G_{\mathbf\theta}(\mathbf{v}_i)(\xi)|^2, 
\end{equation}
where $u_i(\xi)$ is the ground truth and $N$ denotes the total number of functions in the branch network. The weights associated with each sample in Eq. \eqref{eq:loss_func_deeponet} are denoted by $w_i$, which are assumed to be unity in the simplest case. Examples described in Sections \ref{example:darcy_in_complex} 
and \ref{example:flow_in_cavity} are solved with the architecture of conventional data-driven DeepONet. 

During the optimization process, some query points must be penalized more than others in order to satisfy constraints (initial condition, boundary condition). In such cases properly-designed non-uniform training point weights can improve the accuracy. These penalizing parameters can be manually modulated but are often a tedious procedure or should be decided adaptively during the training of the DeepONet \cite{mcclenny2020self,kontolati2022influence}. These parameters in the loss function can be updated by gradient descent side-by-side with the network parameters. The modified loss function is defined as:
\begin{equation}
    \mathcal L(\bm \theta, \bm \lambda) = \frac{1}{N}\sum_{i = 1}^{N} (\bm \lambda)^2|u_i (\xi)- \mathcal G_{\mathbf\theta}(\mathbf{v}_i)(\xi)|^2,
\end{equation}
where $\bm \lambda = \{\lambda_1, \lambda_2, \cdots \lambda_j\}$ are $j$ self-adaptive parameters, each associated with an evaluation point, $\xi_j$. These parameters are constrained to increase monotonically and are always positive. Typically, in a neural network, we minimize the loss function with respect to the network parameters, $\boldsymbol{\theta}$. However, in this approach, we additionally maximize the loss function with respect to the trainable hyper-parameters using a gradient descent/ascent procedure. The modified objective function is defined as:
\begin{equation}
    \min_{\bm \theta} \max_{\bm \lambda}\mathcal L(\bm \theta, \bm \lambda).
\end{equation}
The self-adaptive weights are updated using the gradient descent method, such that
\begin{equation}
    \bm \lambda^{k+1} = \lambda^{k} + \eta_{\lambda}\nabla_{\bm \lambda}\mathcal L(\bm \theta, \bm \lambda),
\end{equation}
where $n_{\bm \lambda}$ is the learning rate of the self-adaptive weights. Implementing self-adaptive weights in \cite{kontolati2022influence} has considerably improved the accuracy prediction of discontinuities or non-smooth features in the solution. Following the inception of DeepONet in 2019 \cite{lu2021_deeponet}, a number of enhancements to traditional architecture were developed to offer inductive bias and speed up training. In subsequent sections, we present several extensions of DeepONet.

\subsection{Feature expansion in DeepONet}
\label{subsec:feature_expansion_deeponet}
In this section, we lay emphasis on learning the operator from paired data using any available prior knowledge of the underlying system. This knowledge can then be directly encapsulated into DeepONet by changing the architecture of the trunk or branch network, which is problem specific. 

\paragraph{Feature expansion in the trunk network:} Additional information of the output function is included as features in the trunk network. For example, in \cite{kontolati2022influence}, we incorporate the knowledge of the dynamics of decaying evolution of the Brusselator reaction-diffusion system by employing a trigonometric feature expansion of the temporal input of the trunk network. Additionally, historical information about dynamical systems could also be used as features in the trunk.

\paragraph{Feature expansion in the branch net:} To incorporate information that has both spatial and temporal dependencies, we use an additional input function in the branch network. In the regularized cavity flow problem presented in \cite{lu2022comprehensive}, an additional trigonometric input function was included in the branch network to incorporate periodic boundary conditions. Similarly, an additional input function is included in \cite{kissas2022learning}, which is created by averaging a feature embedded in the inputs of the branch network over probability distributions that depend on the corresponding query locations of the output function, employing the kernel-coupled attention mechanism, which allows the operator to accurately model correlations between the query locations of the output functions.

\paragraph{POD modes in the trunk:} The standard DeepONet employs the trunk net to learn the basis of the output function from the data. In this approach, the basis functions are pre-computed by performing proper orthogonal decomposition (POD) on the training data (after the mean has been excluded) and using these basis in place of the trunk net. A DNN is employed in the branch net to learn the POD basis coefficients such that the output can be written as:
\begin{equation}
    \mathcal{G}(\mathbf v_1)(\xi) = \sum_{i=1}^p br_i(\eta) \phi_i(\xi) + \phi_0(\xi),
\end{equation}
where $\phi_0(\xi)$ is the mean function of all $\mathbf u_i(\xi), i =1,\dots,N$ computed from the training dataset, and $\{\phi_1, \phi_2, \dots, \phi_p\}$ are the $p$ precomputed POD modes of $\mathbf u_i(\xi)$. Examples described in Sections \ref{example:darcy_in_complex} and \ref{example:flow_in_cavity} are solved with the architecture of POD-DeepONet. 

Additionally, in \cite{oommen2022learning}, we have used a similar feature expansion in the branch net. The branch takes as input the latent dimension of a convolutional autoencoder to learn the dynamic evolution of a two-phase mixture. In \cite{de2022cost}, the authors have used POD modes in the branch net to learn the operator. This approach is often challenging for discontinuous functions because it is similar to global spectral methods trying to approximate non-smooth functions.

\subsection{Multiple input DeepONet}

\begin{figure}
  \begin{center}
    \includegraphics[width=\textwidth]{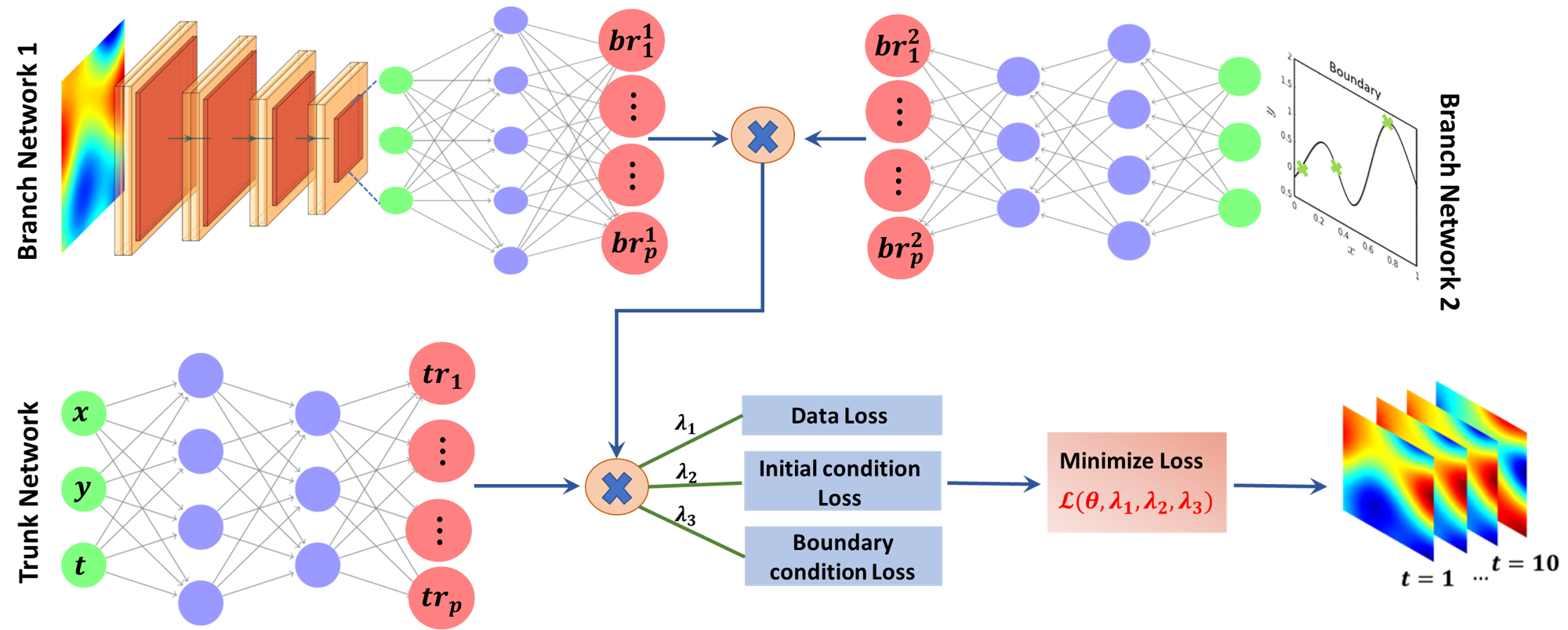}
  \end{center}
  \caption{Schematic representation of a multiple input data-driven DeepONet with self-adaptive weights. The operator takes as inputs the functional field in branch network 1 (employs a CNN), the boundary conditions in branch network 2 (employs an FNN to take as input the values at three sensor locations marked with a green cross mark), and computes the solution for the co-ordinates which are inputs of the trunk network (employs a FNN). The loss function, $\mathcal L$ is the sum of three components; data loss, loss at the initial condition, and the boundary loss. Each of these losses are penalized with penalty parameters, $\lambda_1, \lambda_2, \lambda_3$, respectively. These parameters are updated adaptively along with the weights and biases, $\bm \theta = (W, b)$ of the networks. Hence, $\mathcal L$ is a function of the self-adaptive penalty parameters and the network parameters.}
  \label{fig:don_architecture} 
\end{figure}

The DeepONet, which is based on the universal approximation theorem of operators, is defined for input functions on a single Banach space. To employ DeepONet for realistic setups (multiple input functions) and diverse applications, the multiple input operators theorem was theoretically formulated and proposed in \cite{jin2022mionet}. Such networks are useful for approximating solutions for multiple initial conditions and boundary conditions at the same time. The solution operator for multiple input functions is defined as the tensor product of Banach spaces such that:
\begin{equation}
    \mathcal{G}_{\theta}(\mathbf v, \mathbf w) = \sum_{i=1}^{p} br_i^v\cdot br_i^w\cdot tr_i, 
\end{equation}
where $br_i^v$ and $br_{i}^w$ denote the $i$-th output embedding of the branch networks corresponding to the input functions denoted by $\mathbf v$ and $\mathbf w$, respectively. A schematic representation of this architecture is shown in Figure \ref{fig:don_architecture}. In \cite{goswami2022neural}, a DeepONet framework based on multiple input functions is proposed that encompasses two different DNNs (CNN and FNN) as branch networks. The architecture uses grayscale images of systolic and diastolic geometry (in CNNs) along with patient-specific information (such as hypertension in FNN) to predict the initial distribution and extent of the mechanobiologial insult in a patient with thoracic aortic aneurysm. 

\subsection{Physics-informed DeepONet}\label{sec:pideeponet}

The limitation of purely data-driven approaches is that generalizing the solution requires a large corpus of paired datasets. In many engineering applications, data acquisition is prohibitively expensive, and the amount of available data is typically minimal. As a result, in this ``sparse data" environment, we draw motivation from PINNs to train the DeepONet by directly incorporating known differential equations into the loss function, along with some labeled datasets. The outputs of the DeepONet are differentiable with respect to their input coordinates, thereby allowing the use of automatic differentiation to develop an appropriate regularization mechanism for biasing the target output functions to satisfy the underlying PDE constraints. Keeping in mind the computational cost, for higher dimensional PDEs one could use a finite difference approach or by parametrizing the solution and using a Sobel filter \cite{zhu2019physics} to compute the derivatives.

The hybrid loss function for this framework is defined as:
\begin{equation}\label{eq:phy_deeponet}
    \begin{split}
    &\mathcal L = \mathcal L_{data} + \mathcal L_{physics},\\
    \text{where  } &\mathcal L_{physics} = \mathcal L_{init} + \mathcal L_{bound} + \mathcal L_{pde},\\
    &\mathcal L_{init} = \frac{1}{N_{init}}\sum_{1=1}^{N_{init}}\left[\mathcal G_{\theta}(\mathbf v)(x,y,t_0) - \mathbf u(x,y,t_0)\right]^2,\\
    &\mathcal L_{bound} = \frac{1}{N_{bound}}\sum_{i=1}^{N_{bound}}\left[\mathcal G_{\theta}(\mathbf v)(x_b,y_b,t) - \mathbf u(x_b,y_b,t)\right]^2,\\
    &\mathcal L_{pde} = \frac{1}{N_f}\sum_{i=1}^{N_f}\left[f(\mathcal G_\theta)\right]^2,
    \end{split}
\end{equation}
where $N_{init}$ denotes the number of initial data points, $N_{bound}$ denotes the number of boundary points, and $N_f$ denotes the number of collocation points or integration points on which the PDE is evaluated. Additionally, $f(\mathcal G_{\theta})$ denotes the residual form \cite{wang2021learning} or the variational form \cite{goswami2022physics} of the governing equation. $\mathcal L_{pde}$ acts as an appropriate regularization mechanism for biasing the target output functions to satisfy the underlying PDE constraints, when very few labeled data are available. Examples discussed in Sections \ref{example:fracture} and \ref{example:hetero_porus_media} are solved using the additional physics-informed loss term to obtain the optimized network parameters.

\section{FNO and its extensions}
\label{sec:fno}
\begin{figure}
  \begin{center}
    \includegraphics[width=\textwidth]{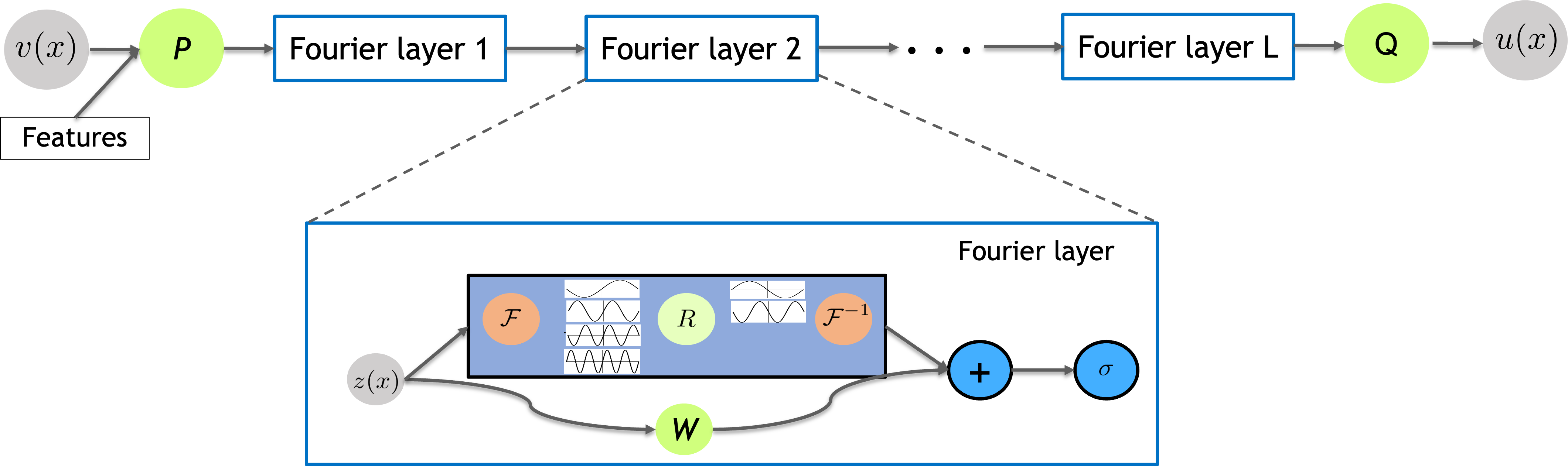}
  \end{center}
  \vspace*{-0.1cm}
  \caption{Schematic representation Fourier neural operator. The input function, $\mathbf v(x)$ is transformed to a higher dimensional representation using a shallow neural network, $\mathcal{P}$ and then operated by a series of $L$ Fourier layers. Within each Fourier layer, a Fourier transform, $\mathcal F$ of the input is obtained to filter out the higher modes using a linear transform, $R$, and then converted back to physical space using inverse Fourier transform, $\mathcal F^{-1}$. Along with the Fourier space transformation, a residual connection with a weight matrix, $\mathbf W$ is applied on the input function and added and is later acted upon by a non-linear activation function. At the end of the Fourier layers, a local transformation $\mathcal{Q}$ is applied by employed a shallow neural network to convert the output space to a dimension of the input grid points.}
  \label{fig:fno} 
\end{figure}

The Fourier neural operator (FNO) is based on replacing the kernel integral operator by a convolution operator defined in Fourier space.  The operator takes as input functions defined on a well-defined equally spaced lattice grid and outputs the field of interest on the same grid points. The network parameters are defined and learned in the Fourier space rather than in the physical space, \textit{i.e.,} the coefficients of the Fourier series of the output function are learned from the data. A schematic representation of the FNO is shown in Figure \ref{fig:fno}, which can be viewed as a DeepONet with a convolution neural network in the branch net to approximate the input functions and Fourier basis functions in the trunk net. In particular, the network has three components: First, the input function $\mathbf{v}(x)$ is lifted to a higher dimensional representation $\hb(x,0)$, through a lifting layer, $\mathcal{P}$, which is often parameterized by a linear transformation or a shallow neural network.  Then, the neural network architecture is formulated in an iterative manner: $\hb(x,0)\rightarrow \hb(x,1)\rightarrow\hb(x,2)\rightarrow \cdots \rightarrow \hb(x,L)$, where $\hb(x,j)$, $j=0,\cdots,L$, is a sequence of functions representing the values of the architecture at each layer. Each layer is defined as a nonlinear operator via the action of the sum of Fourier transformations and a bias function:
\begin{align}
\nonumber\hb(x,j+1)=&\mcL_j^{FNO}[\hb(x,j)]\\
:=&\sigma\left(W_j\hb(x,j)+\mathcal{F}^{-1}[R_j\cdot \mathcal{F}[\hb(\cdot,j)]](x)+ \cb_j\right).\label{eq:FNO}
\end{align}
Here, we use $\sigma$ to denote the activation function. $W_j$, $\cb_j$ and $R_j$ are trainable parameters for the $j$-th layer, such that each layer has different parameters (i.e., different kernels, weights and biases). Lastly, the output $\mathbf{u}(x)$ is obtained by projecting $\hb(x,L)$ through a local transformation operator layer, $\mathbf{Q}$. In Section \ref{example:darcy_in_square}, we demonstrate the performance of conventional data-driven FNO on learning the solution operator for PDEs. 

\subsection{Feature expansion in FNO}
\label{subsec:feature_expansion_fno}

One important requirement of FNO is that the input function be defined on a lattice grid, which makes FNO often difficult to apply for problems where the input function defined on few points (such as the boundary condition, initial condition) to be mapped to the solution over the whole domain. Additionally, if the problem domain is defined on an unstructured mesh like in complex geometries, implementing the conventional FNO architecture is a challenge. To address these issues, several feature expansions were proposed in \cite{lu2022comprehensive}.

\paragraph{dFNO+:} This feature is implemented for problem setup where the domain of the input function is different from the domain of the output function. For example, if we want to map the initial condition to the spatial and temporal evolution of the solution, we define the mapping as
\begin{equation}
    \mathcal G: \mathbf v(x,0) \rightarrow \mathbf{v}(x,t), 
\end{equation}
where $\mathbf{v}(x,0)$ defines the initial condition, and $\mathbf{v}(x,t)$ is the evolved dynamics. In such cases, the input space is defined on the spatial domain and so it is difficult to map the output to the spatial and the temporal domain simultaneously. To this end, we propose two approaches. In the first approach, we define a new input function, $\tilde{\mathbf{v}}(x,t)$, which has an additional temporal component and is defined as $\tilde {\mathbf{v}}(x,t) = \mathbf{v}(x)$. 
As our second approach, we propose to define the output space employing a recurrent neural network such that the solution operator is decomposed into a series of operators and the solution of each time step is obtained iteratively using a time marching scheme such that $\mathcal G: \mathbf v(x,t) \rightarrow \mathbf{v}(x,t+\Delta t)$.
Alternatively, the input space could be defined as a subset of the output space, while attempting to map the boundary condition to the solution defined over the entire domain. In such cases, the input function can be padded with zeros for the domain's interior points and then considered as input to the neural operator.

\paragraph{gFNO+:} FNO employs discrete Fast Fourier Transform (FFT), which necessitates the definition of the input and output functions on a Cartesian domain with a lattice grid mesh. However, for problems defined on complex real-life geometry, an unstructured mesh is typically used, requiring us to deal with two issues: (1) non-Cartesian domain and (2) non-lattice mesh. 
To handle issues associated with input and output function defined on non-Cartesian domain, we define a bounding box, and project the input and the output space by ``nearest neighbor" to maintain continuity at the boundaries. Alternatively, for issues associated with the non-lattice mesh, we perform interpolation between the unstructured mesh and a lattice grid mesh. The examples described in Sections \ref{example:darcy_in_complex} and \ref{example:flow_in_cavity} are solved with the combination of dFNO+ and gFNO+. 

\paragraph{Wavelet Neural Operators:} In \cite{tripura2022wavelet}, the Wavelet Neural Operator (WNO) was proposed, which learns the network parameters in the wavelet space that are both frequency and spatial localized, thereby can learn the patterns in the images and/or signals more effectively. Specifically, the Fourier integral of FNO was replaced by wavelet integrals for capturing the spatial behaviour of a signal or for studying the system under complex boundary conditions. It was shown that WNO can handle domains with both smooth and complex geometries, and it was applied in learning solution operators for highly nonlinear family of PDEs with discontinuities and abrupt changes in the solution domain and the boundary.

\subsection{{Implicit FNO}} 

\begin{figure}
  \begin{center}
    \includegraphics[width=0.8\textwidth]{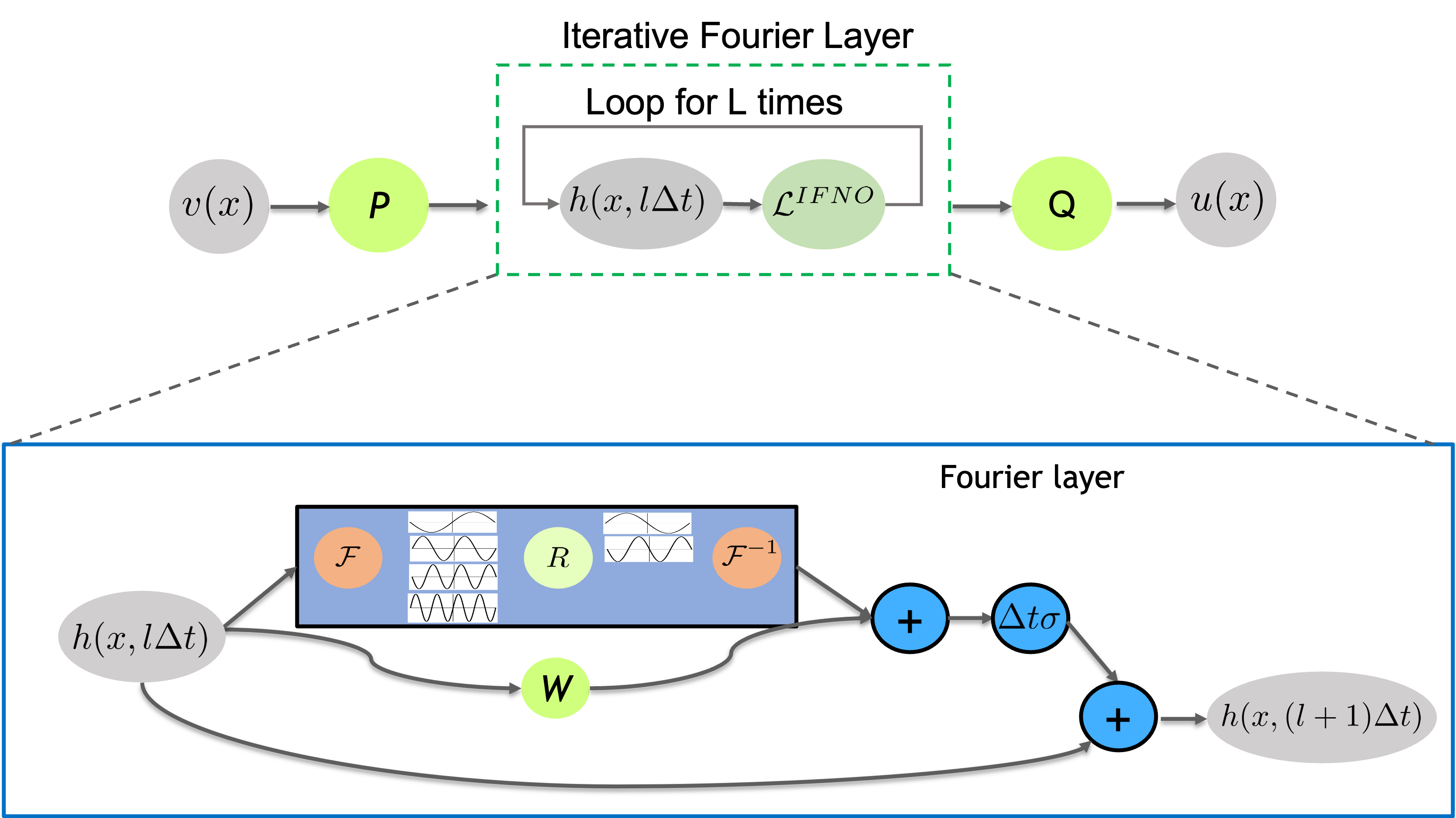}
  \end{center}
  \vspace*{-0.1cm}
  \caption{A schematic representation for the implicit Fourier Neural Operator (IFNO), which enhances the vanilla FNOs architecture with reduced memory cost and improved stability in the deep layer limit. This architecture also employs the lifting layer, $\mathcal{P}$, and the projection layer, $\mathcal{Q}$, as in the original FNOs, and proposes a modified model for the Fourier layer. Within each Fourier layer, the number of layers is identified with the number of time steps in a time-discretization scheme, and the increments between layers are parameterized via the action of the sum of Fourier space transformation and the local linear transformation, in a manner that all Fourier layers share the same set of trainable parameters. As such, the iterative layers can be interpreted as a discretized autonomous integral differential equation.}
   \vspace*{-0.5cm}
  \label{fig:ifno} 
\end{figure}

In the vanilla FNO, to guarantee the universal approximation property different trainable parameters are employed for each Fourier layer \cite{kovachki2021universal}. Hence, the number of trainable parameters increases as the network gets deeper, which makes the training process of the FNO more challenging and potentially prone to over-fitting. In \cite{you2022nonlocal} (see Figure \ref{fig:loss_2ddarcy_s31} of Section \ref{example:darcy_in_square}), it was found that when the network gets deeper, the training error decreases in the FNO while the test error becomes much larger than the training error, indicating that the network is overfitting the training data. Furthermore, if one further increases the number of hidden layer $L$, training the FNOs becomes challenging due to the vanishing gradient phenomenon.

To improve the neural network's stability performance in the deep layer limit, in \cite{you2022learning} You et al. proposed to model the PDE solutions of unknown governing laws as the implicit mappings between given loading/boundary conditions and the resultant solution, with the neural network serving as a surrogate for the solution operator. Based on this idea, the implicit FNO (IFNO) architecture was developed, which can be interpreted as a data-driven surrogate of the fixed point procedure, in the sense that the increment of fixed point iterations is modeled as an autonomous operator between layers. As illustrated in Figure \ref{fig:ifno}, in IFNOs the same parameter set is employed for each iterative layer, with the layer update writes:
\begin{align}
\nonumber\hb(x,t+\Delta t)=&\mcL^{IFNO}[\hb(x,t)]\\
:=&\hb(x,t)+ {\Delta t}\sigma\left(W\hb(x,t)+\mathcal{F}^{-1}[R\cdot \mathcal{F}[\hb(\cdot,t)]](x)+ \cb\right).\label{eq:IFNO}
\end{align}
Here, the trainable parameters $W$, $R$ and $\cb$ are taken to be layer-independent, so the number of trainable parameters does not increase with the number of layers, alleviating the major bottleneck of the over-fitting issue encountered by the original FNO in Eq. \eqref{eq:FNO}. Moreover, this feature also enables the straightforward application of the shallow-to-deep initialization technique \cite{ruthotto2019deep}. In fact, the index of integral layers is identified with the number of time steps in a time-discretization scheme. By dividing both sides of Eq.~\eqref{eq:IFNO} by $\Delta t$, the term $(\hb(\cdot,t+\Delta t)-\hb(\cdot,t))/\Delta t$ corresponds to the discretization of a first order derivative, and Eq. \eqref{eq:IFNO} can be interpreted as a nonlinear differential equation in the limit of deep layers, i.e. as $\Delta t\to 0$. Thus, the optimal parameters ($W$, $R$ and $\cb$) of a shallow network can be interpolated and reused in a deeper one as initial guesses.

In IFNO, a forward pass through a very deep network is analogous to obtaining the PDE solution as an implicit problem, and the universal approximation capability is guaranteed as far as there exists a convergent fixed point equation. Since the proposed architecture is built as a modification of the FNO, it also parameterizes the integral kernel directly in the Fourier space and utilizes the fast Fourier transformation (FFT) to efficiently evaluate the integral operator. Hence, IFNO inherits the advantages of FNO on resolution independence and efficiency, while demonstrates not only enhanced stability but also improved accuracy in the deep network limit. In Section \ref{example:tissue}, we demonstrate the performance of IFNO on biological tissue
modeling based on digital image correlation (DIC) measurements.

\subsection{Physics-Informed FNO}\label{sec:pifno}

Motivated by physics-informed DeepONet (PI-DeepONet) proposed in \cite{goswami2022physics} and \cite{wang2021learning}, the physics-informed FNO (PINO) was proposed in \cite{li2021physics,konuk2021physics} as an integration of operator learning and physics-informed settings. PINO reduces the labelled dataset requirement for training the neural operator and helps in faster convergence of the solution. In this setting, it is important to note that unlike DeepONet, FNO outputs the solution on a grid that employs FFT. Hence, the gradients of the output function with respect to the inputs space cannot be computed using the automatic differentiation library commonly employed in machine learning algorithms. To this end, the following approaches can be used to explicitly define the gradients:
\begin{enumerate}
    \item Using the conventional numerical gradients such as finite difference and Fourier gradient. These approaches require either a fine discretization (for finite-difference), else the numerical error would be magnified, or would require smooth and uniform grid (spectral methods).
    \item Applying automatic differentiation of the sum of the Fourier coefficient at every spatial location (without doing the inverse FFT) and the value of a query function defined as an interpolation or a low-rank integral operator \cite{kovachki2021neural}.
    \item Explicitly defining the gradients on the Fourier space and applying the chain rule to compute the required quantities.
\end{enumerate}
The authors of \cite{li2021physics} have computed the exact gradient by defining the gradient on the Fourier space. Additionally, the linear mapping that arises due to the residual connection with the weight matrix, $\mathbf W$ shown in Figure \ref{fig:fno}, is interpolated using the Fourier method. To optimize the network parameters, the loss function defined in Eq. \eqref{eq:phy_deeponet} is minimized. 

Besides the scenarios where full physics constraints, such as the known governing equations, are provided, in many real-world modeling tasks only partial physics laws are available. To improve the learning efficacy on such problems, in \cite{you2022physics} the authors proposed to impose partial physics knowledge via a soft penalty constraint to the neural operator. 
In particular, an IFNO was built to model the heterogeneous material responses from the DIC displacement tracking measurements of multiple bi-axial stretching protocols on a porcine tricuspid valve anterior leaflet. Both the constitutive model and the material micro-structure are unknown, and hence there is no known governing law that can be imposed. The authors then proposed to infuse the no-permanent-set assumption to guide the training and prediction of the neural operators. In other words, when the specimen is at rest, one should observe zero displacement field in the specimen. Specifically, a hybrid loss function $\mcL=\mcL_{data}+\lambda\mcL_{physics}$ is employed, with the same data-driven loss as defined in Eq.  \eqref{eq:phy_deeponet} and the physics-informed loss defined as a penalization term:
$$\mcL_{physics}:=\left[\mathcal G_{\theta}(\mathbf 0)(x) \right]^2.$$
Here, $\mathbf 0(x)$ denotes the input function valued zero everywhere, and $\lambda>0$ is a tunable hyper-parameter to balance the data-driven loss and the physics-informed loss. This method was shown to improve the extrapolative performance of IFNO in the small deformation regime.

\section{Graph Neural Operators}
\label{sec:gno}
Machine learning using graphs is becoming increasingly popular mainly because of its ability to capture the graph (non-local) structure of the data. The concept of every node of a graph consisting of information from every other node is quite enriching. Machine learning tasks using a graph neural network can be of different types. For example, it can be a node level prediction (where one wants to determine what is the attribute or properties of a particular node), edge level prediction (whether or not an edge exits between given nodes) or it may be a full graph prediction (where one wants to classify the whole graph). 

\subsection{Graph Neural Networks}

\begin{figure}
    \centering
    \includegraphics[width=0.8\textwidth]{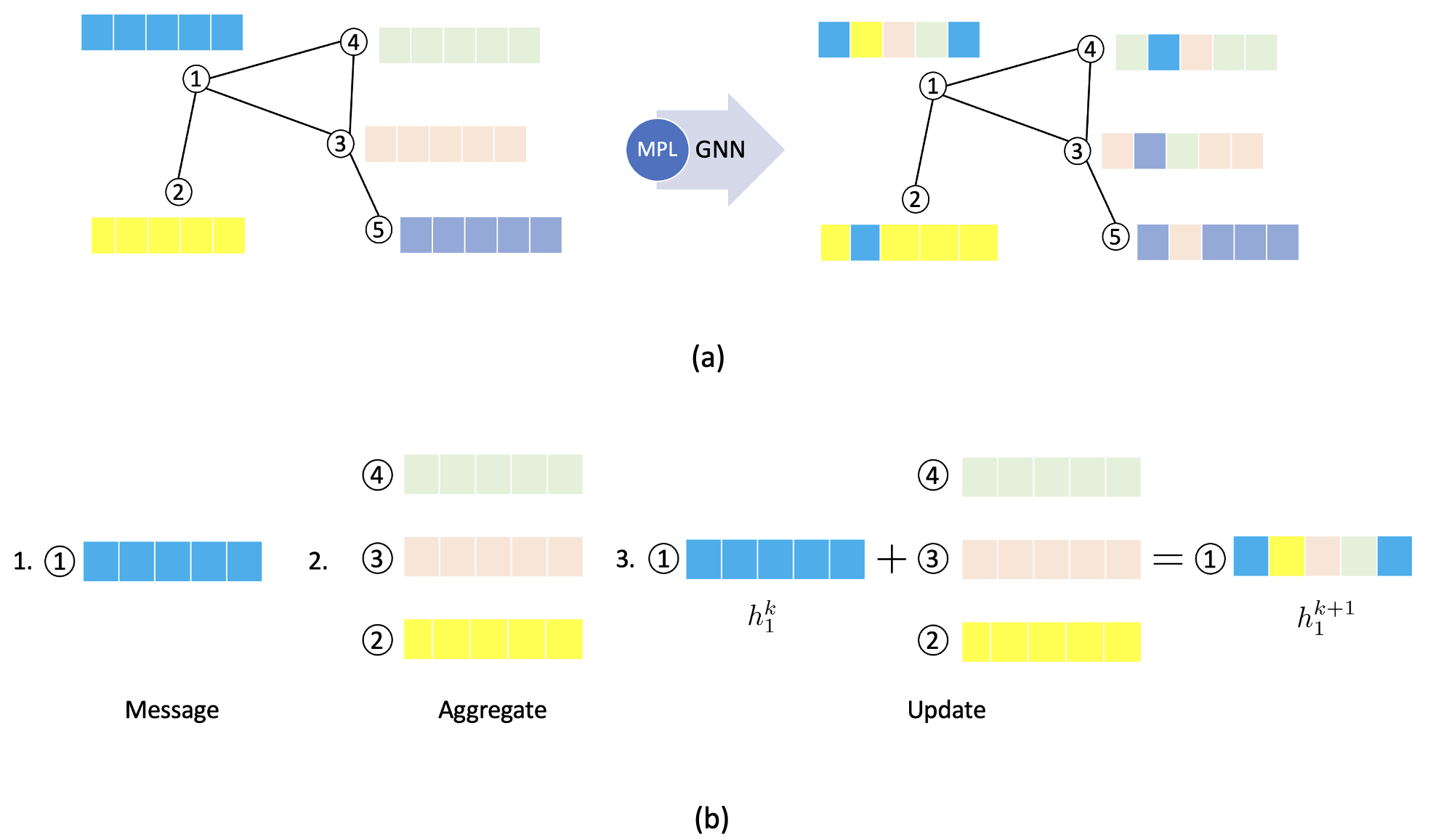}
    \caption{Illustration depicting a message passing layer for a GNN, where 5(a) shows a general message passing layer effect on the node features of a graph and 5(b) shows the steps inside it.}
    \label{fig:GNN}
\end{figure}

A graph neural network (GNN) is a transformation, which can be optimized based on all attributes of the graph (nodes, edges, global-context). This transformation has to preserve the graph symmetry (permutation invariance). In GNNs, the message passing layer (MPL) plays a key role and forms the core component. The set of operations that happens in message passing layers can be summarized in three major steps: (1) the Message step where information is gathered for the each of the node and its edges; (2) the Aggregate step where the information from all the neighbouring nodes is gathered for each node; and (3) the Update step where the information from the previous two steps are combined to update the state of each node. 
Mathematically, the three steps can be expressed with the equation
\begin{equation}
\hb_{i}^{(k+1)}=\operatorname{update}^{(k)}\left(\hb_{i}^{(k)}, \operatorname{aggregate}^{(k)}\left(\left\{\hb_{j}^{(k)}, \forall j \in \mathcal{N}(i)\right\}\right)\right),
\end{equation}where $\hb_{i}^{(k)}$ is the current state $k$, for node $i$, and  $\mathcal{N}(i)$ represents the neighbourhood of node $i$ which consists of the nodes that have a direct edge connection with $i$. Here, $\operatorname{update}$ and $\operatorname{aggregate}$ can be defined as different kind of functions based on specific learning tasks, such as $\operatorname{mean}$, $\operatorname{max}$, $\operatorname{normalized}$ $\operatorname{sum}$, a Multi Layer Perceptron (MLP), and Recurrent Neural Network (RNN), just to name a few. Up-to-date, the two major GNNs that are widely used are Graph Convolution Network \cite{kipf2016semi} and Graph Attention Network \cite{velivckovic2017graph}.

\subsection{Integral Neural Operators through Graph Kernel Learning}

Here, we introduce the graph kernel network (GKN) approach and its variations, which can be interpreted as a continuous version of a GNN as well as an integral neural operator. As a motivating example, let us consider a PDE of the form:
$$
\begin{gathered}
\left(\mathcal{L}_{a} u\right)(x)=v(x), \quad x \in \Omega\;, \\
u(x)=0, \quad x \in \partial \Omega,
\end{gathered}
$$
where $a(x)$ is the parameter field, $v(x)$ is the loading term which acts as the input function in the solution operator and $u(x)$ is the PDE solution which can be seen as the output function. We can define the Green's function $G: \Omega \times \Omega \rightarrow \mathbb{R}$ as the unique solution to the problem under relatively general constraints on $\mathcal{L}_{a}$, such that
$$
\mathcal{L}_{a} G(x, \cdot)=\delta_{x}\;,
$$
where $\delta_{x}$ is the delta measure on $\mathbb{R}^{d}$ centered at $x$. Because $G(x,y)$ is dependent on the coefficient $a$, we will refer to it as $G_{a}$ from here on. Then the ground-truth solution operator, $\mathcal{G}$, can be expressed as an integral operator of the Green's function:
$$
u(x)=\mathcal{G}(v)(x):=\int_{\Omega} G_{a}(x, y) v(y) d y
$$
When $\mathcal{L}_a$ is uniformly elliptic, for example, the Green's function is generally continuous at positions $x \neq y$, and hence using a neural network $\kappa$ to model the kernel becomes intuitive. In particular, the model writes: 
\begin{equation}\label{eqn:greens}
u(x)=\int_{\Omega} \kappa(x, y, a(x), a(y)) v(y) d y,
\end{equation}
with $\kappa$ being a shallow neural network taking $x$, $y$, $a(x)$ and $a(y)$ as its inputs. Based on this idea, two graph-based neural operators, namely, the graph kernel network \cite{Li_GKN} and the non-local kernel network \cite{you2022nonlocal} were constructed and will be entailed below. Here, both graph-based neural operators were constructed with the data-driven loss only, and hence the physics is introduced through data. We also point out that the ideas of imposing full or partial physics in Sections \ref{sec:pideeponet} and \ref{sec:pifno} can be easily applied for these graph-based neural operators as well.

\paragraph{Graph Kernel Networks (GKNs).} The idea of GKNs comes from parameterizing the Green's function in an iterative architecture \cite{Li_GKN}. As an integral neural network similar to the FNOs, GKNs are also composed by a lifting layer $\mathcal{P}$, iterative kernel integration layers, and a projection layer $\mathcal{Q}$. While the lifting layer and the projection layer share the same architecture as the FNO, it is assumed that the iterative kernel integration part is invariant across layers, 
with the update of each layer network given via the action of the summation of a non-local integral operator Eq. \eqref{eqn:greens} and a linear operator:
\begin{align}
\nonumber\hb(x,j+1)=&\mathcal{L}^{GKN}[\hb(x,j)]\\
:=&\sigma\left(W\hb(x,j)+\int_\Omega \kappa(x,y,a(x),a(y);\theta)\hb(y,j) dy + \cb\right).\label{eq:gkn}
\end{align}
Similar as in FNOs, in GKNs the nodes within each layer are treated as a continuum, so each layer representation can be seen by a function of the continuum set of nodes $D\subset\real^d$.  
$\kappa\in\real^{s\times s}$ is a tensor kernel function that takes the form of a (usually shallow) NN whose parameters $\theta$ are to be learned through training. Different from the setting in FNOs, in GKNs the parameters $W$, $\cb$ and $\theta$ are layer-independent. As such, the GKN resembles the original ResNet block \cite{He2016Resnet}, where the usual discrete affine transformation is substituted by a continuous integral operator. 
In practice, the interaction range of kernel $\kappa(x,y,a(x),a(y);\theta)$ is often chosen based on the known information about the true kernel of the application, or based on the computational efficiency needs.  When taking the interaction range as $\Omega$, i.e., every point in the whole domain has impact on $x$, then, the model will be more expressive but computationally expensive. On the other hand, one can restrict the interaction range as the ball with radius $r$ centered at $x$, i.e., $B_r(x)$, for efficiency purposes, keeping in mind that this choice might compromise the accuracy $B_r(x)$.

In \cite{Li_GKN}, the integral in Eq. \eqref{eq:gkn} is realized through a message passing graph neural network architecture. In particular, the physical domain $\Omega$ is assumed to be discretized as a set of points $\chi:=\{x_1,\cdots, x_J\}\subset \Omega$. Then, these points are treated as the nodes of a weighted, directed graph, such that an edge $\{i,j\}$ presents when the representation of point $x_j$ has impact to the representation of point $x_i$, i.e., $\kappa(x_i,x_j,a(x_i),a(x_j)))\neq 0$. Denoting $N(x)$ as the neighborhood of each point $x\in\chi$ according to the graph, with the message passing algorithm of \cite{gilmer2017neural} the integral operator in Eq. \eqref{eq:gkn} is implemented as an averaging aggregation of messages:
\begin{align}
\hb(x,j+1)=&\sigma\left(W\hb(x,j)+\dfrac{1}{|N(x)|}\sum_{y\in N(x)} \kappa(x,y,a(x),a(y);\theta)\hb(y,j) + \cb\right),\label{eq:gkn_disc}
\end{align}
where $|N(x)|$ represents the total number of points in $N(x)$. When taking the interaction range as the whole domain, i.e., $N(x_i)=\chi$ for all $x_i$, the corresponding graph of Eq. \eqref{eq:gkn_disc} will be fully-connected and the number of edges scales like $O(J^2)$, which makes the GKNs generally much more expensive than some other neural operators, say, FNOs. To accelerate the computation of GKNs, several techniques were implemented. In \cite{Li_GKN}, the Nystr\"{o}m approximation method is considered, which samples uniformly at random the points of $N(x)$, so as to reduce the complexity of computation when calculating the integral. In \cite{li2020multipole}, the multipole graph neural operator (MGNO) is proposed. By unifying a multi-resolution matrix factorization of the kernel with GNNs, MGNOs capture global properties of the PDE solution operator with a linear time-complexity. As another approach to improve the learning efficiency, in \cite{gupta2021multiwavelet} the authors proposed to approximate the kernel of the integral operator through a better basis representation by using the multiwavelet transform.

\begin{figure}
    \centering
    \includegraphics[width=0.48\textwidth]{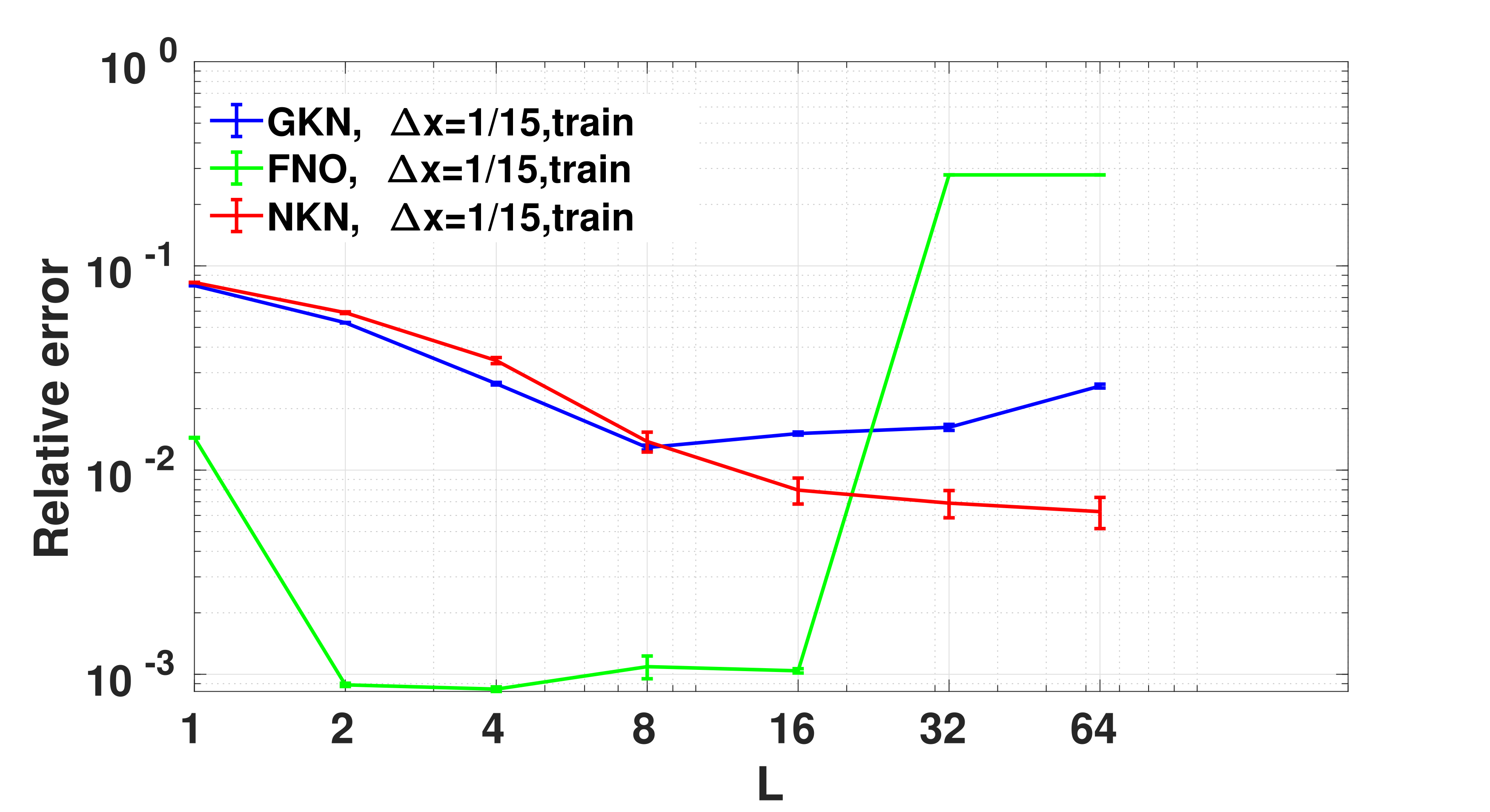}
    \includegraphics[width=0.48\textwidth]{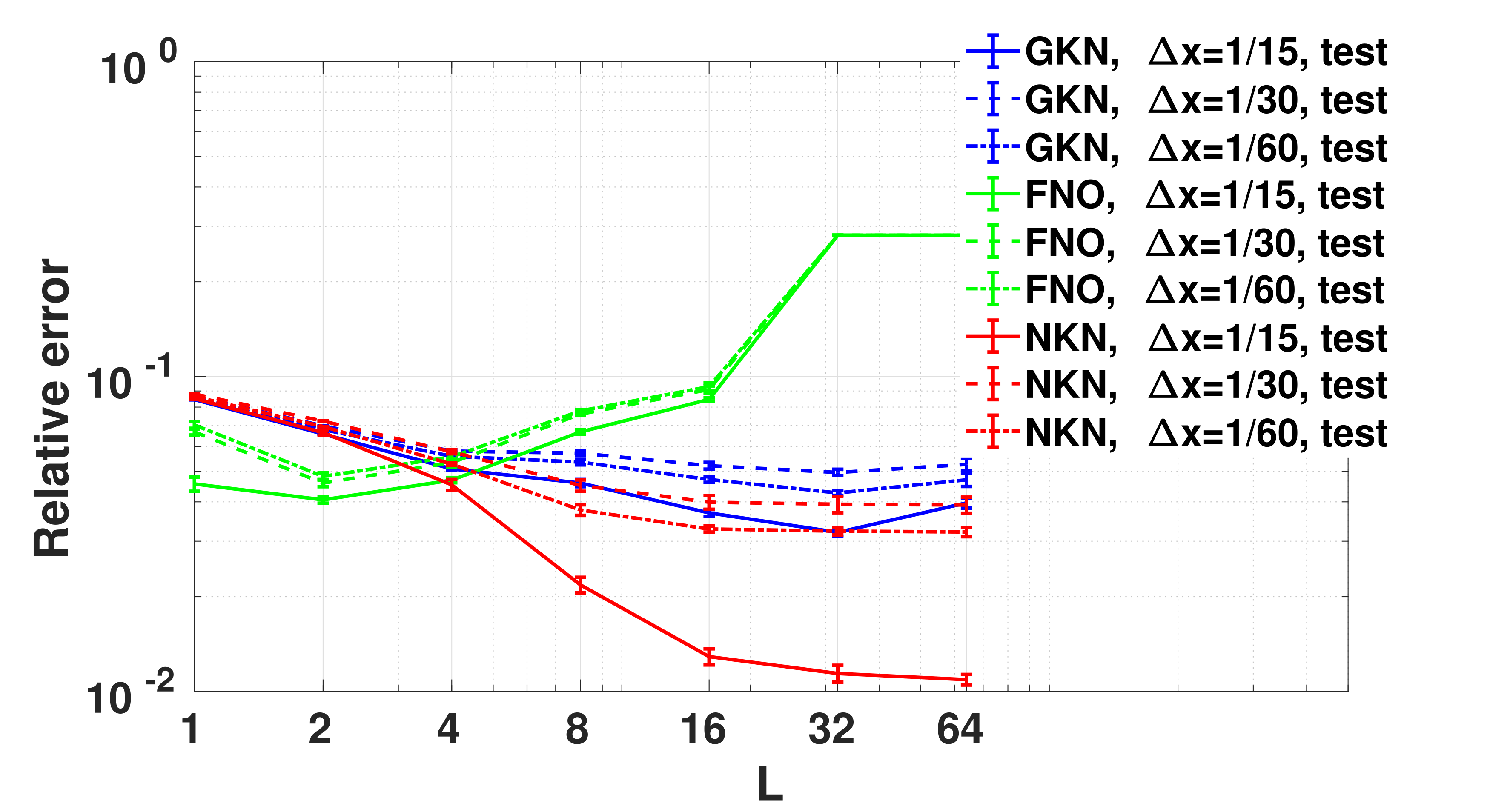}
    \caption{{\bf 2D Darcy flow in a square domain.} Demonstration of the stability performance of three integral neural operators. Comparison of relative mean squared errors from GKNs, FNOs, and NKNs when using the training set with grid size $\Delta x=1/15$ \cite{you2022nonlocal}. Error bars represent standard errors over 5 simulations. Left: errors on the training dataset. Right: errors on the test dataset with different resolutions: $\Delta x=1/15$, $\Delta x=1/30$, and $\Delta x=1/60$. Detailed experimental settings and further numerical results are provided in Section \ref{sec:darcy}.}
    \label{fig:loss_2ddarcy_s31}
\end{figure}

\paragraph{Non-local Kernel Networks (NKNs).} Although GKNs are resolution-independent, in \cite{you2022nonlocal} it was found that GKNs might become unstable when increasing the number of iterative kernel integration layers, $L$. As shown in Figure \ref{fig:loss_2ddarcy_s31}, when $L$ increases, then either there is no gain in accuracy or increasing values of the loss function occur in GKNs. To obtain a reliable deep kernel network, which can handle more complicated and general learning tasks, in \cite{you2022nonlocal} the non-local kernel network (NKN) was proposed. NKNs share the same architecture of the lifting and projection layers as the GKNs, FNOs and IFNOs, but propose to use an alternative form of the iterative integration layer. Similar to the IFNOs, in NKNs the index of integral layers is also identified with the number of time steps in a time-discretization scheme, by letting $t=j\Delta t$ and given the $(j+1)$-th network layer presentation by
\begin{align}
\nonumber\hb(x,t+\Delta t)=&\mathcal{L}^{NKN}[\hb(x,t)]\\
\nonumber:=&\hb(x,t)+ {\Delta t}\left(\int_\Omega \kappa(x,y,a(x),a(y);\theta)(\hb(y,t)-\hb(x,t)) dy\right.\\
&\left.-\beta(x;w)\hb(x,t)+\cb\right).\label{eq:NKN}
\end{align}
Here, the kernel tensor function $\kappa$ is modeled by a NN parameterized by $\theta$, and a reaction term, the tensor function $\beta$, is modeled by another NN parameterized by $\zeta$. Both $\kappa$ and $\beta$ are usually shallow NNs, such as the multilayer perceptron (MLP), with the parameters $\theta$ and $\zeta$ to be learned together with the biases, $\cb$. One can see that the integral operator on the right-hand side of Eq. \eqref{eq:NKN} can be interpreted as a non-local Laplacian operator:
$$\mathcal{L}^{diff}_\kappa[\hb]:=\int_{\Omega} \kappa(x,y,a(x),a(y))(\hb(y,t)-\hb(x,t)) dy.$$
Therefore, the NKN architecture can indeed be interpreted as a nonlocal diffusion reaction equation in the limit of deep layers, i.e., as $\Delta t\to 0$:
$$\dfrac{\partial \hb}{\partial t}(x,t) -\mathcal{L}^{diff}_\kappa[\hb](x,t) +\beta(x)\hb(x,t)=\cb.$$
The stability of NKNs was analyzed via non-local vector calculus, showing that when the kernel function $\kappa$ is square integrable and non-negative, and the reaction parameter function $\beta$ is positive and bounded, the learnt non-local operator is positive definite and the network is stable in the limit of deep layers. In \cite{you2022nonlocal}, when applied to the Poison equation solution learning task it was found that the NKNs' amplification matrix is positive definite, while the GKNs' matrix exhibits negative eigenvalues, indicating that instabilities might occur. In Section \ref{sec:darcy}, we demonstrate the performance of NKNs, GKNs, and the vanilla FNOs, on learning the solution operator for the 2D Darcy's equation.

\begin{remark}
Here, we would like to point out that a similar idea of considering the correspondence between the stable architecture and the stable PDEs was also recognized recently in Graph Neural Diffusion (GRAND) \cite{chamberlain2021grand}. In GRAND, the authors interpreted GNN architectures from a mathematical framework by different choices of the form of diffusion equation and discretization schemes. It was showed that more advanced and stable numerical schemes such as Runge-Kutta and implicit schemes would help to improve the performance, and amount to larger multi-hop diffusion operators in the design of deep GNN architectures.
\end{remark}

\section{Neural Operator Theory}

Beyond the pioneering work of Chen \& Chen \cite{chen_chen_approximation} on the universal approximation theory of operators for a single layer, other works have appeared only recently for deep neural networks. The first theoretical work that extended the Chen \& Chen theorem to deep neural networks was in \cite{lu2021_deeponet}. The paper by Deng et al. \cite{deng2022approximation} considers advection-diffusion equation, including nonlinear cases. The authors have shown that DeepONet has exponential approximation rates in the linear case. Moreover, they demonstrated that by emulating numerical methods, DeepONet has algebraic convergence with respect to the network size. In the paper by Lanthaler et al. \cite{lanthaler2022error}, the authors have extended the universal approximation theorem in \cite{chen_chen_approximation} and have removed the continuity and compactness assumptions. Also, they have provided an upper bound for the DeepONet error by decomposing it into three parts: encoding error, approximation error and  reconstruction error. They have also proven lower bounds on the reconstruction error by utilizing optimal errors for projections on finite-dimensional affine subspaces of separable Hilbert spaces. They have used this to prove the two-sided bounds on the DeepONet error. This construction also allows them to infer the size of trunk net needed to approximate the eigen-functions of the covariance operator in-order to obtain optimal reconstruction errors. In \cite{marcati2021exponential}, the authors have shown that for linear second order PDEs with non-homogeneous coefficients and source terms, DeepONet has exponential expression rates for the coefficient-to-solution operators in the $H^1$ norm. Additionally, their results also show that neural networks can emulate accurately the (discrete) solution map of Galerkin methods for the elliptic PDEs mentioned above with numerical integration. They have also proven that the DeepONet architecture has size $\mathcal{O}(|log(\epsilon)|^{\kappa})$ for any $\kappa> 0$ depending on the physical space dimension. In paper \cite{yu2021arbitrary}, the authors  have shown that for non-polynomial activation functions, an operator with neural network of width five is arbitrarily close to any given continuous nonlinear operator. They have also shown the theoretical advantages of depth by constructing operator ReLU neural networks of depth $2k^{3}+8$ with constant width, which they compare with other operator ReLU neural network of depth k. 

In paper \cite{kovachki2021universal}, the authors have shown that FNOs are universal, i.e., they can approximate any continuous operator to desired accuracy. However, they have also shown that in the worst case, the size of a FNO can grow super-exponentially in terms of the desired error for approximating a general Lipschitz continuous operator. They have proved rigorously that there exists a $\psi$-FNO, which can approximate the underlying nonlinear operators efficiently for Darcy flow and the  incompressible Navier-Stokes equations. In paper \cite{you2022learning}, You et al. have shown that the IFNO is a universal solution finding operator, in the sense that it can approximate a fixed point method to a desired accuracy. 

Apart from the data driven approaches discussed above, the paper \cite{de2022generi} has presented error bounds for physics-informed operator learning for both DeepONets and FNOs. 
Finally, in the paper \cite{garcia2020error}, the authors have proposed a general framework to analyze the rates of spectral convergence for a large family of graph Laplacians. They established a convergence rate of $O\left(\left(\frac{\log n}{n}\right)^{\frac{1}{2 m}}\right)$. Also, in \cite{Li_GKN}, Li et al have shown that the Graph Kernel Network approach has competitive approximation accuracy to classical and deep learning methods.

\section{Applications}
\label{sec:application}

This section is divided into two parts that present several numerical examples that were solved using either data-driven training of the neural operators (Section \ref{sec:DD-Training}) or physics-informed training (Section \ref{sec:PI_training}) to evaluate the performance of the neural operators discussed above. To assess the efficacy of neural operators, we compute the relative $L_2$ error of the predictions. Each example includes information about the data generation, network architecture, activation function, and optimizer used.

\subsection{Data driven neural operators}
\label{sec:DD-Training}
In this section, we present three problems to show the implementation of data-driven neural operators. 

\subsubsection{Darcy flow in a square domain: }\label{sec:darcy}
\label{example:darcy_in_square}As the first benchmark example, we consider the modeling problem of two-dimensional sub-surface flows through a porous medium with heterogeneous permeability field. In this example, we compare the results obtained using data-driven FNOs, GKNs, and NKNs. The high-fidelity synthetic simulation data for this example are described by the Darcy flow, which was first proposed in \cite{Li_GKN}, and later considered in a series of neural operator studies \cite{Li_GKN,li2020multipole,Li_FNO,lu2022comprehensive,you2022nonlocal,you2022learning}. The governing differential equation is defined as:
\begin{equation}\label{eq:darcy_PDE}
\begin{split}
    -\nabla \cdot (K(\bm x)\nabla u(\bm x)) &= f,\;\;\; \bm x = (x,y),\\
    \text{subjected to}\;\;\;u_D(\bm x) &= 0, \;\;\; \forall \;\; \bm x \in \partial \Omega,
\end{split}
\end{equation}
where $K$ is the conductivity field, $u$ is the hydraulic head, and $f$ is a source term. In this context, the goal is to learn the solution operator of the Darcy's equation and compute the solution field $u(\xb)$.

\begin{figure}
    \centering
    \includegraphics[width=\textwidth]{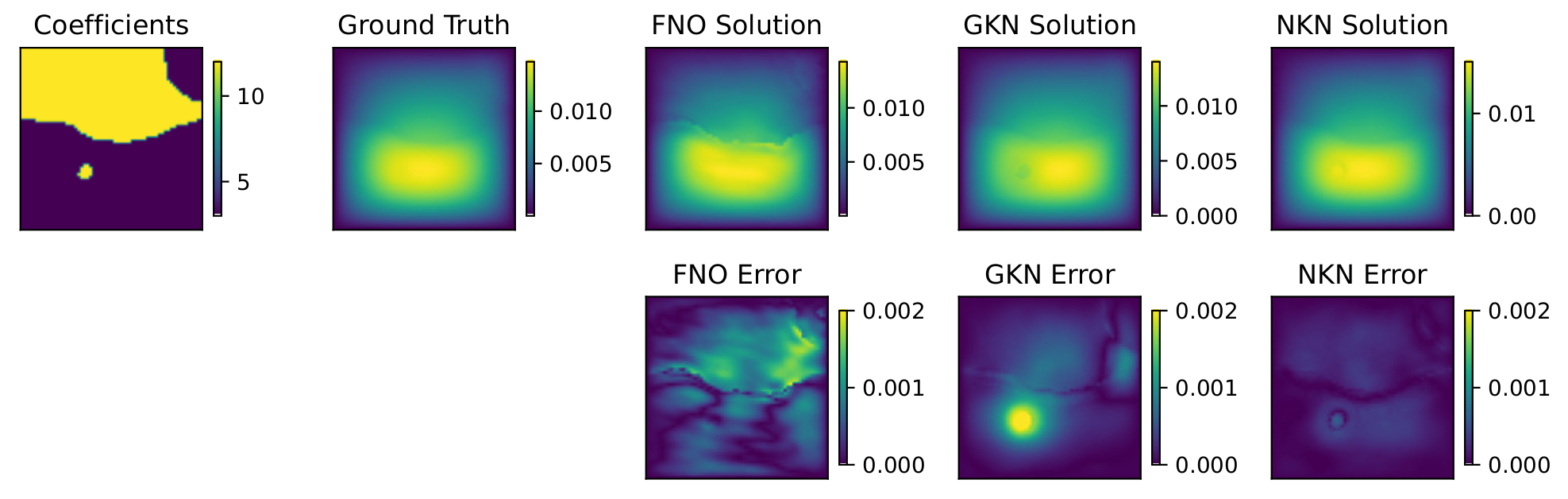}
    \caption{{\bf 2D Darcy flow in a square domain.} A visualization of 16-layer FNO, GKN, and NKN performance on a instance of conductivity parameter field $K(\xb)$, when using (normalized) ``coarse'' training dataset ($\Delta x=1/15$) and test on the ``finer'' dataset ($\Delta x=1/60$).}
    \label{fig:s61_test}
\end{figure}

In this example, we consider a square physical domain $\Omega=[0,1]^2$, a fixed source field $f(\xb)=1$ and Dirichlet boundary condition $u_D(\xb)=0$. The aim is to obtain the solution field $u(\xb)$ for each realization of conductivity field $K(\xb)$. That means, the neural operators are employed to learn the mapping from $K(\xb)$ to $u(\xb)$. As standard simulations of subsurface flow, the permeability $K(\xb)$ is modeled as a two-valued piecewise constant function with random geometry such that the two values have a ratio of 4. Specifically, $140$ samples of $K(\xb)$ were generated according to $K\sim \psi_{\#}\mcN(0,(-\Delta+9I)^{-2})$, where $\psi$ takes a value of 12 on the positive part of the real line and a value of 3 on the negative. $100$ samples are employed for the purpose of training, while the rest forms the test dataset. Different resolutions of data sets are down-sampled from a $241\times 241$ grid solution generated by using a second-order finite difference scheme, as provided in \cite{Li_GKN}. The corresponding data can be found at \cite{gkn_github}. With the purpose of testing generalization properties with respect to resolution, three data sets with different resolutions are considered here: a ``coarse'' data set with grid size {$\Delta x=1/15$}, a ``fine'' data set with grid size {$\Delta x=1/30$}, and a ``finer'' data set with grid size {$\Delta x=1/60$}. Here, the coarse set is employed for the purpose of training, and the performance of the resultant neural operators was tested on data sets with all resolutions. 

\paragraph{Results:} We report the training/test errors of FNO, GKN, and NKN in Figure \ref{fig:loss_2ddarcy_s31} and the plot of solutions in one representative test sample in the ``finer'' data set in Figure \ref{fig:s61_test}, where all neural operators are with $L=16$ layers. One can observe that all solutions obtained with NKN are visually consistent with the ground-truth solutions, while GKN loses accuracy near the material interfaces. FNO results are off in a even larger regions.  These results provide further qualitative demonstration of the loss of accuracy in FNOs and GKNs from the previous sections. In this case, the averaged relative test errors from GKN, FNO and NKN are 4.71\%, 9.29\%, and 3.28\%, respectively. The results for this problem have been adapted from \cite{you2022nonlocal}.

\subsubsection{Darcy flow in a complex domain: }
\label{example:darcy_in_complex}
We now further consider Darcy flows in a triangular domain with a vertical notch to map the boundary condition to the hydraulic head, $u(\bm x)$, with $K(x, y) = 0.1$, and $f = -1$. In this illustration, we assess the outcomes of data-driven DeepONet, dgFNO+, and POD-DeepONet.

The following Gaussian process is employed to generate samples of boundary conditions for each boundary in the triangular domain:
\begin{equation}\label{eq:Gaussian_boundary}
\begin{split}
    u_D(x) &\sim \mathcal{GP}(0, \mathcal{K}((x,y), (x',y'))), \\
    \mathcal{K}(\bm x, \bm x') &= \exp[-\frac{(\bm x - \bm x')^2}{2 l^2}], ~l = 0.2, \text{  and  }\bm x, \bm x' \in [0, 1].
\end{split}
\end{equation}
Specifically, $1084$ nodes are employed in the numerical solver to discretize the geometry, and $2000$ different boundary conditions are generated of which $1900$ realizations are used for training, and the remaining are used to test the accuracy of the neural operators for unseen cases. We generate the paired dataset by solving Eq. \eqref{eq:darcy_PDE} using the \emph{MATLAB Partial Differential Equation Toolbox}.

\begin{figure}[htbp]
\centering
\includegraphics[width=0.9\textwidth]{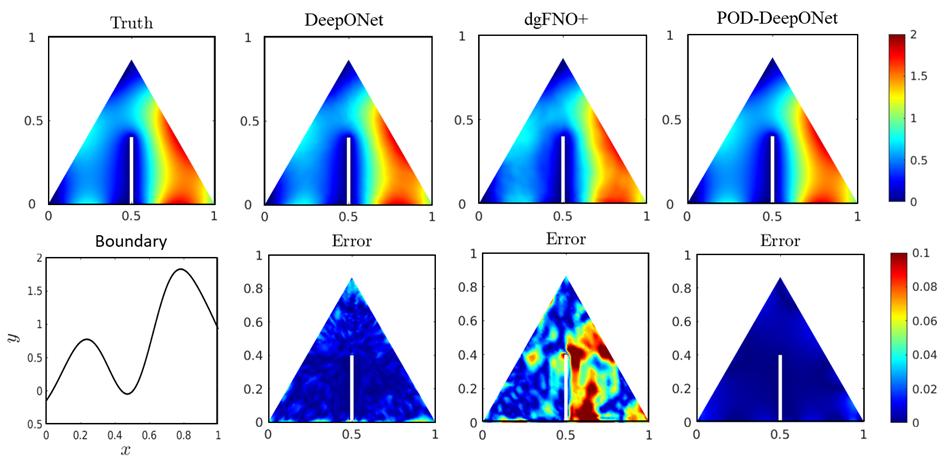}
\caption{\textbf{Darcy flow in a triangular domain with a notch.} For a representative boundary condition, the hydraulic head is obtained using DeepONet, dgFNO+, and POD-DeepONet. The prediction errors for the three operator networks are shown against the respective plots. The ground truth is simulated using the {\emph{PDE Toolbox}} in Matlab. The predicted solutions and the ground truth share the same colorbar, while the errors corresponding to each of the neural operators are plotted on the same colorbar. The results are adopted from \cite{lu2022comprehensive}.}
\label{fig:darcy_triagular_notch}
\end{figure}

For DeepONet training, we have employed FNNs as both branch and trunk networks with $3$ hidden layers consisting of $128$ neurons in each layer for the trunk network, while $2$ hidden layers of $128$ neurons in the branch network. The network parameters are optimized using the Adam optimizer and the \textit{ReLU} activation function is used. Additionally, we now employ POD-deepONet to approximate the hydraulic head. To this end, we employ an FNN architecture with $3$ hidden layers of $128$ neurons in the branch network, and the trunk consists of $32$ dominant POD modes. To train the Fourier neural operator, we have mapped the triangular domain with a notch to a bounded square domain. To map the boundary conditions to the solution of the hydraulic head, we implement a combination of feature expansions (dgFNO +) discussed in Section \ref{subsec:feature_expansion_fno}. The network architecture for FNO employs a shallow neural network, $\mathbf P$ of $32$ neurons, followed by $4$ Fourier layers that filter out $8$ lowest modes. The last transformation, $\mathbf Q$ is defined using an FNN with $2$ layers of $128$ neurons each, and employs the activation function \textit{ReLU}. 
\paragraph{Results:} A relative $L_2$ error of $1.0\%$, $2.02\%$, and $7.12\%$ is obtained using POD-DeepONet, DeepONet and dgFNO+. A representative case is shown in Figure \ref{fig:darcy_triagular_notch}, where the predicted results and the errors corresponding to DeepONet, dgFNO+, and POD-DeepONet are shown for a specific boundary.

\subsubsection{Flow in a cavity: }
\label{example:flow_in_cavity}
In this example, we consider a two-dimensional lid-driven flow in a square cavity (i.e., $\Omega= [0, 1]^2$), to compare the accuracy of the network to accurately predict the velocity of the flow field for a given boundary condition using data driven DeepONet, dFNO+ and POD-DeepONet. The flow field can be described by the incompressible Navier-Stokes equations as:
\begin{align}
    \nabla \cdot \mathbf{u} &= 0,\\ 
    \partial_t \mathbf{u} + \mathbf{u} \cdot \nabla \mathbf{u} &= -\nabla P + \nu \nabla^2 \mathbf{u}, 
\end{align}
where $\mathbf{u} = (u, v)$, where $u$ is the velocity in the $x-$ direction and $v$ is the velocity in the $y-$ direction, $P$ denotes the pressure, and $\nu$ is the kinematic viscosity of the fluid. We consider a case with different boundary conditions for the upper wall. In particular, the boundary conditions are expressed as:
\begin{align}
    u = U\left(1 - \frac{\cosh[r(x - \frac{L}{2})]}{\cosh(\frac{rL}{2})}\right), ~ v = 0,
\end{align}
where $U$, $r$ and $L$ are constants. In this setup, $r = 10$, $L = 1$ is the length of the cavity. In addition, the remaining walls are assumed to be stationary. The aforementioned equations are then solved using the lattice Boltzmann method \cite{meng2015multiple} to generate the training data. Interested readers could find more details on data generation in \cite{lu2022comprehensive}. To generate the labeled dataset, we generate 100 velocity flow fields at various Reynolds numbers, $Re = [100,\;2080]$, with a step size of 20. A total of $100$ labeled datasets of the flow field were simulated, $90$ of which were used for training the neural operator and $10$ were used as unseen cases to test the accuracy of the trained network. The operator network maps the upper wall boundary condition to the converged velocity field. 
The DeepONet architecture uses a convolution neural network in the branch net and a FNN with $3$ hidden layers of $128$ neurons in the trunk net. The hyperbolic tangent activation function \textit{tanh} is used for this problem. Next, we use POD-DeepONet with the same branch net architecture and $6$ modes to approximate the flow fields. 

\begin{figure}[htbp]
\centering
\includegraphics[width=0.9\textwidth]{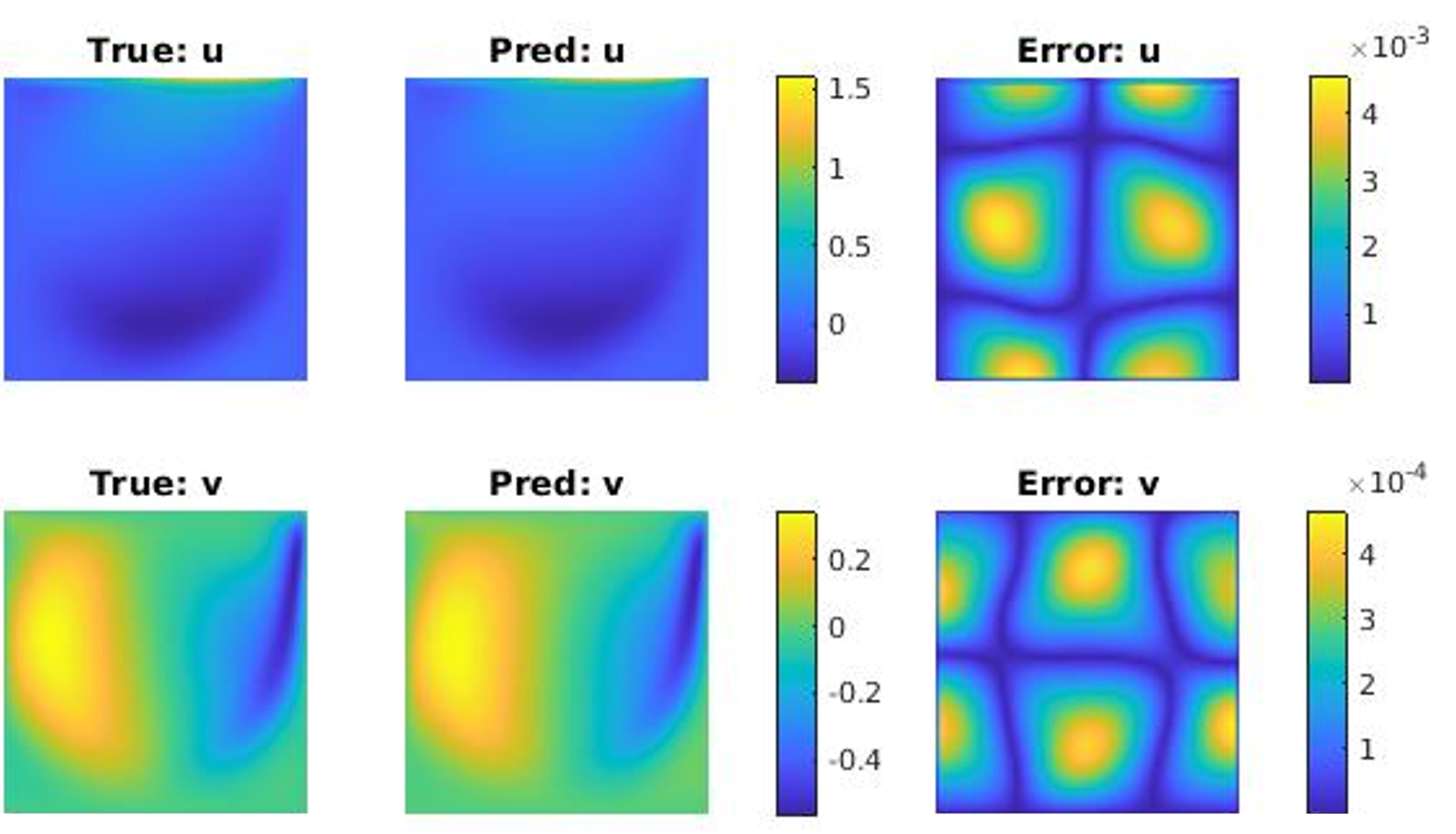}
\caption{\textbf{Steady cavity flow described by Navier-Strokes Equation.} A representative case from the test data set depicting the flow fields, $u$ and $v$ in $x-$ and $y-$ directions, respectively and the associated errors obtained using POD-DeepONet. The predicted solutions and the ground truth share the same colorbar, while the errors corresponding to each of the neural operators are plotted on the same colorbar.}
\label{fig:cavity}
\end{figure}

\paragraph{Results: }The relative $L_2$ error is reported as $1.20\%$, $0.33\%$ and $0.63\%$ for predicting flow fields for unseen boundary conditions using DeepONet, POD-DeepONet and dFNO+. A representative case is shown in Figure \ref{fig:cavity}, where the predicted results and errors corresponding to POD-DeepONet are shown for a specific boundary. The results for this problem have been adapted from \cite{lu2022comprehensive}.

\subsection{Physics-informed neural operators}
\label{sec:PI_training}
In this section, we consider an additional loss term, $\mathcal L_{physics}$ as a regularization term in the loss function, to optimize the hyperparameters of the neural operators along with few labelled datasets (see Sections \ref{example:fracture} and \ref{example:hetero_porus_media}) and/or improve the neural operators' generalizability (see Section \ref{example:tissue}).

\subsubsection{Brittle fracture in a plate loaded in shear: }
\label{example:fracture}
In this example, we aim to model the final crack path, given any initial location of the crack on a unit square plate with a horizontal crack at the middle height from the left outer edge, fixed at left and bottom edge, is subjected to shear loading on the top edge. This example illustrate the benefits of using a physics-driven loss comparing the results obtained using PI-DeepONet and data-driven DeepONet. We use the phase field approach \cite{goswami2019adaptive} to model the crack in the domain. The material properties considered are: $\nu = $ 121.15 kN/mm$^{2}$, $\mu = $ 80.77 kN/mm$^{2}$, as Lam\'e's first and second parameters, respectively, and the critical energy release rate, $G_c = 2.7 \times 10^{-3}$ kN/mm. The boundary conditions of the setup is denoted as:
\begin{equation}
    u(x,0) = v(x,0) = 0, \;\;\; u(x,1)= \Delta u,
\end{equation}
where $u$ and $v$ are the solutions of the elastic field in \textit{x} and \textit{y}-axis, respectively and $\Delta u$ is the incrementally applied shear displacement on the top edge of the plate. To train the PI-DeepONet using the variational form, we minimize the total energy of the system, which is defined as:
\begin{equation}\label{eq:totalenergy}
    \begin{split}
        &\mathcal{E} = \Psi_e + \Psi_c,\\
\text{where}\;\;\;&\Psi_e =  \int_{\Omega}f_e(\mathbf{x}) d\xb,\;\;\; f_e(\mathbf{x}) =  (1-\phi)^2\psi_{e}^+\left(\bm{\epsilon}\right) + \psi_{e}^-\left(\bm{\epsilon}\right),\\
&\Psi_c =  \int\limits_\Omega f_c(\mathbf{x}) d\xb,\;\; f_c(\mathbf{x}) =  \frac{G_c}{2l_0} \left(\phi^2 +l_0^2 |\nabla\phi|^2 \right) - (1-\phi)^2H(\mathbf{x},t),
    \end{split}    
\end{equation}
where $\Psi_e$ is the stored elastic strain energy, $\Psi_c$ is the fracture energy, $l_0$ is the length scale parameter that controls the diffusion of the crack, $\psi_{e}^{+}$ and $\psi_{e}^{-}$ are the tensile and the compressive components of the strain energies densities obtained by the spectral decomposition of the strain tensor, and $H(\mathbf{x},t)$ is the strain-history functional. In this example, we have used a hybrid loss function to train the network parameters. The training samples are obtained ($n = 11$) for different initial crack lengths, $l_c \in [0.2, 0.7]$ in steps of $0.05$. For the network architecture, the branch net and the trunk net are 4-layers fully-connected neural networks with $[100, 50, 50, 50]$ neurons, respectively. Once the solution is evaluated at the sampled points, the outputs for the elastic field are modified to exactly satisfy the Dirichlet boundary conditions, as:
\begin{equation}\label{eq:shear_boundary_auto}
\begin{split}
    \mathcal G_{\bm{\theta}}^{u} &= [y(1-y)]\hat{\mathcal G}_{\bm{\theta}}^{u} + y \Delta u,\\
    \mathcal G_{\bm{\theta}}^{u} &= [y(y-1)]\times [x(x-1)] \hat{\mathcal G}_{\bm{\theta}}^{v}, 
\end{split}
\end{equation}
where $\hat{\mathcal G}_{\bm{\theta}}^{u}$ and $\hat{\mathcal G}_{\bm{\theta}}^{v}$ are obtained from the DeepONet. The conventional data-driven DeepONet is trained with the same $11$ samples, keeping the network architecture of the branch net and the trunk net exactly the same. The synthetic data to train the network is generated using the codes developed in \cite{goswami2020Aadaptive}. To improve the accuracy, the training samples are increased to $43$.

\begin{figure}[t!]
    \centering
    \subfigure{
    \includegraphics[trim=62 32 8 15,clip,width = 0.65\textwidth]{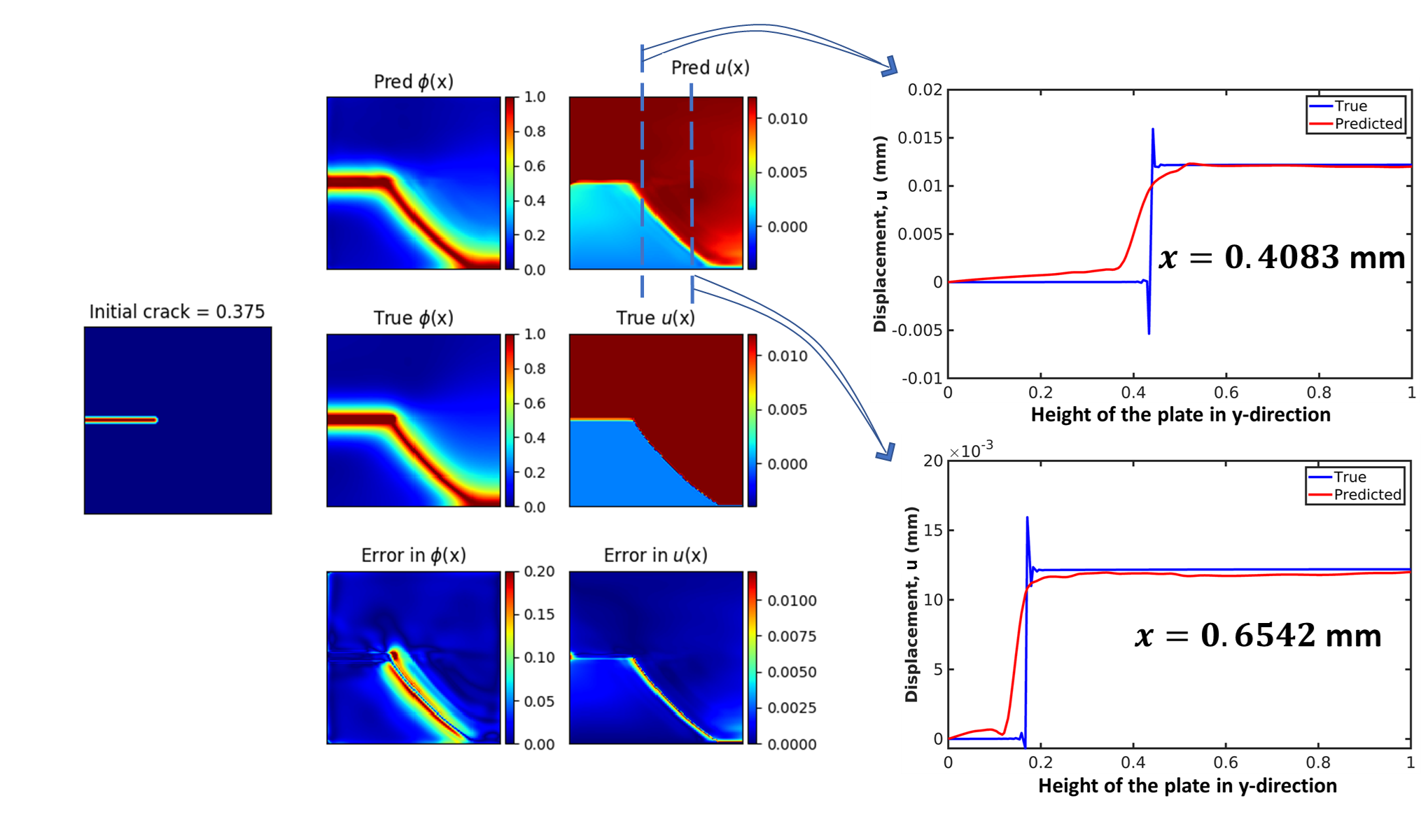}}
    \caption{Shear failure: the PI-DeepONet is trained with $11$ crack lengths to predict the final damage path for any crack length, when the height of the crack is fixed at the centre of the left edge. The plot is for $l_c = 0.375$ mm, where $\Delta u = 0.220$ mm. The predicted displacement in $x$-direction is plotted for two locations along the $x$-axis and is compared with ground truth to show the accuracy of the prediction. The results are adopted from \cite{goswami2022physics}.}
    \label{fig:shear_Xtest_final_DD}
\end{figure}

\paragraph{Results: }A prediction error of $2.16\%$ on $\phi$ is reported when PI-DeepONet was employed. Additionally, predictions of data-driven DeepONet have an error of $26.2\%$ and $3.12\%$ for $\phi$ when trained with $11$ samples and $43$ samples, respectively. Figure \ref{fig:shear_Xtest_final_DD} presents the plots of the predicted solutions for $l_c = 0.375$ mm is presented, which is obtained using PI-DeepONet. The  results for the data driven DeepONet suggest that it is unable to capture the crack diffusion phenomenon and also it cannot generalize to complex fracture phenomenon with limited data-sets.

\subsubsection{Flow in heterogeneous porous media: }
\label{example:hetero_porus_media}
We consider a two-dimensional flow through heterogeneous porous media, which is governed by Eq. \eqref{eq:darcy_PDE} using PI-DeepONet, which uses no labelled datasets and is trained using the governing physics of the problem. In this example, we aim to learn the operator such that:
\begin{equation}
    \mathcal G_{\bm{\theta}}:K(\bm x) \rightarrow h(\bm x), 
\end{equation}
where $K(\bm x)$ is spatially varying hydraulic conductivity, and $h(\bm x)$ is the hydraulic head. The setup is of a unit square plate with a discontinuity of $5\times10^{-3}$ mm. For generating multiple conductivity field for training the neural operator, we describe the conductivity field, $K(\bm x)$, as a stochastic process. In particular, we take $K(\bm x) = \exp(F(\bm x))$, with $F(\bm x)$ denoting a truncated Karhunen-Lo\`eve (KL) expansion for a certain Gaussian process, which is a finite-dimensional random variable. The DeepONet is trained using the variational formulation of the governing equation, and without any labelled input-output datasets. The optimization problem can be defined as:
\begin{equation}\label{eq:Darcy_energy}
\begin{split}
    &\text{Minimize:}\;\;\;\;\; \mathcal{E} = \Psi_h,\\
    &\text{subject to:}\;\;\;\; h(\mathbf{x}) = 0 \text{ on } \partial \Omega_{D},\\
\end{split}
\end{equation}
where $\partial \Omega_D$ represents the boundary of the domain and 
\begin{equation}\label{eq:energyterms_Darcy}
       \Psi_h = \frac{1}{2}\int_{\Omega}K(\bm x)|\nabla h(\bm x)|^2 \;\; d\xb- \int_{\Omega}h(\bm x)\;\; d\xb.
\end{equation}
The network architecture of the branch and the trunk networks are two separate 6-hidden layers FNN with 32 neurons per hidden layer.

\begin{figure}[!htbp]
    \centering
    \subfigure[]{
    \includegraphics[trim=20 0 40 80,clip,width= \textwidth]{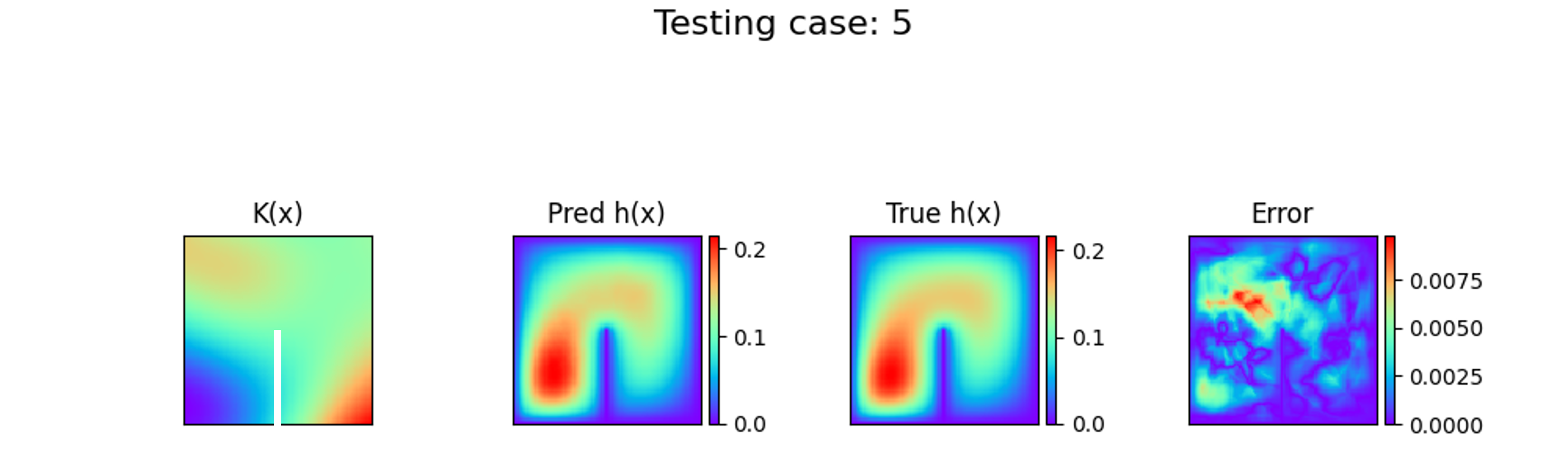}}
    \subfigure[]{
    \includegraphics[width = 0.55\textwidth]{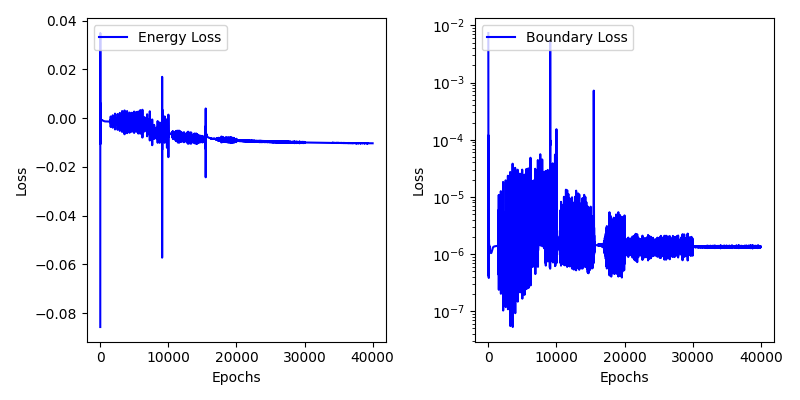}}
    \caption{Flow in heterogeneous porous media: (a) The predicted $h(\bm x)$ for a given conductivity field, $K(\bm x)$ (plotted on log scale) is shown for a representative samples. True $h(\bm x)$ represents the ground truth and is the simulated solution using the Matlab PDE toolbox. The difference between the predicted $h(\bm x)$ and the ground truth is shown in the error plot. (b) The plots show the loss trajectory of the two components of the loss function. The plot on the left shows the decrease in energy of the domain with respect to the number of epochs, while the plot on the right shows the boundary loss term. The results are adopted from \cite{goswami2022physics}.}
    \label{fig:darcy_plots_test}
\end{figure}

\paragraph{Results: }The trained PI-DeepONet yields a predictive error of $3.12\%$. The loss trajectory is shown in Figure \ref{fig:darcy_plots_test}(b). The prediction of $h(\bm x)$ for a representative sample of $K(\bm x)$, using PI-DeepONet is shown in Figure \ref{fig:darcy_plots_test}(a). It is interesting to note that we have tried to solve the problem by minimizing the residual \cite{wang2021learning}. However, the residual based DeepONet is not able to approximate the solution of $h(\bm x)$ for a given $K(\bm x)$.

\subsubsection{Biological tissue modeling from experimental measurements: } 
\label{example:tissue}

\begin{figure}[!ht]
\centering
\subfigure{\includegraphics[width=1.0\columnwidth]{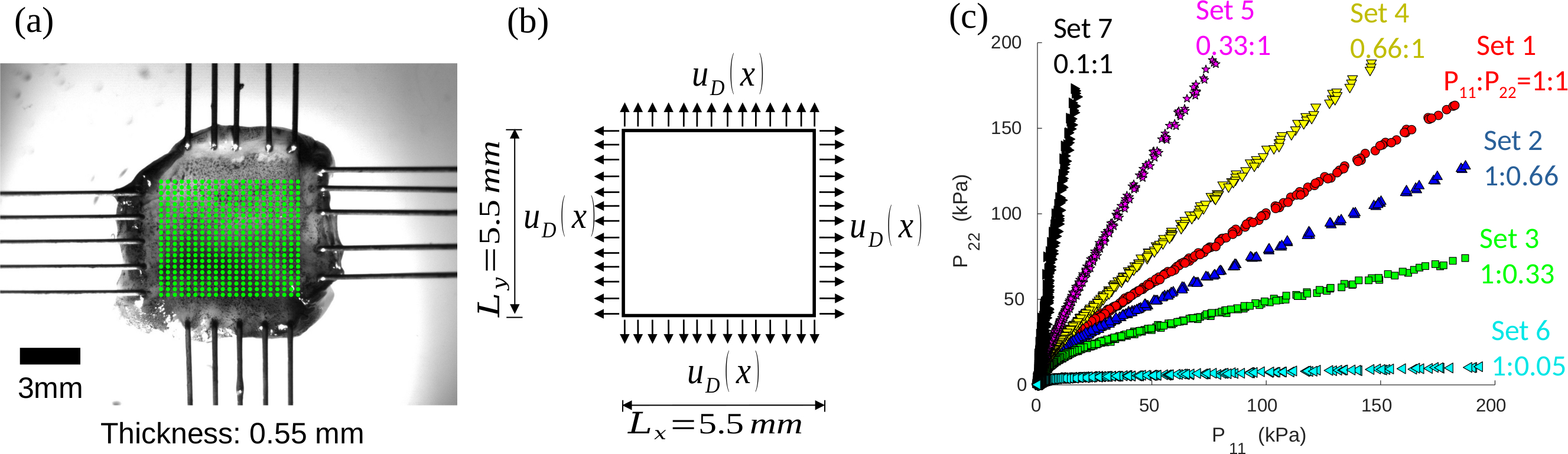}}
\caption{{\bf Problem and experimental setups for the biological tissue modeling example.} (a) An image of the speckle-patterned porcine tricuspid valve anterior leaflet (TVAL) specimen subject to biaxial stretching, with the DIC tracking grid shown in green. (b) Schematic of a specimen subject to Dirichlet-type boundary conditions, so the goal of neural operator learning is to provide a surrogate mapping from the boundary displacement $u_D(\xb)$, $\xb\in\partial\Omega$, to the displacement field $u(\xb)$, $\xb\in\Omega$. (c) Illustration of the seven protocols of the mechanical testing on a representative TVAL specimen. Here,  $P_{11}$ and $P_{22}$ denote the first Piola-Kirchhoff stresses in the $x$- and $y$-directions, respectively.}
\label{fig:dicsetup}
\end{figure}

\begin{figure*}[h!]
    \centering
    \subfigure{\includegraphics[width=1.\textwidth]{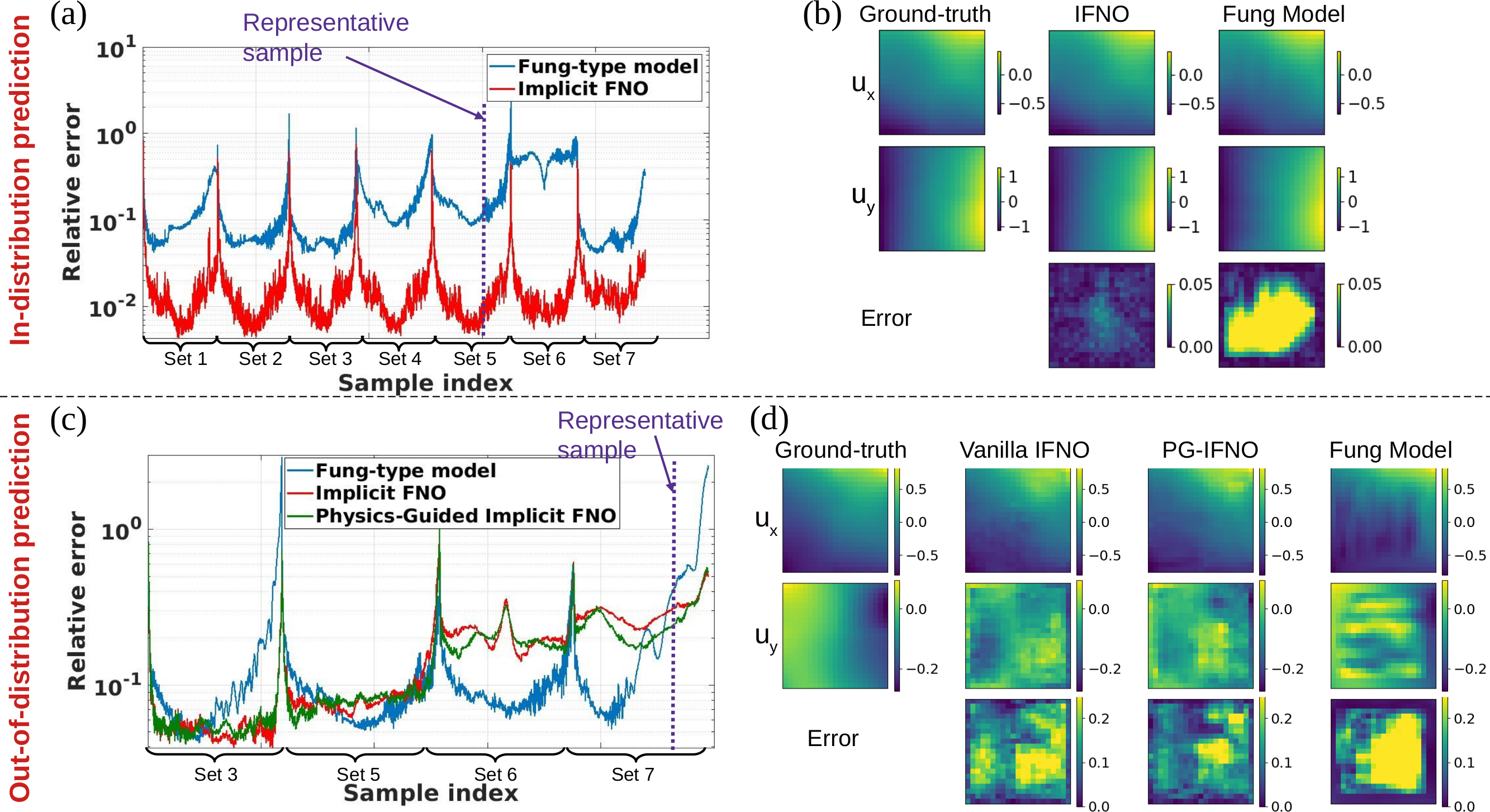}}
    \caption{{\bf Biological tissue modeling from DIC displacement data.} Upper: in-distribution prediction with training set on 83\% of randomly selected samples. (a) Sample-wise error comparison between a conventional constitutive model (the Fung-type model) and the implicit FNO on all biaxial testing protocol sets. (b) Visualization of the Fung-type model fitting and implicit FNO performances on a representative test sample. Bottom: out-of-distribution prediction on the small deformation regime, by training each model on protocol sets 1, 2, and 4 then testing on the rest of sets. (c) Sample-wise error comparison between the Fung-type model, the original implicit FNO, and the physics-guided implicit FNO on testing protocol sets. (d) Visualization of the Fung-type model fitting and physics-guided implicit FNO performances on a representative test sample.}
    \label{fig:unsmoothdata_study1}
\end{figure*}

To demonstrate the applicability and generalizability of neural operators on modeling complex systems from noisy real-world data, in \cite{you2022physics} the implicit FNO was employed to predict the material displacement field of biological tissue without postulating a specific constitutive model form nor possessing knowledges on the material microstructure, from digital image correlation (DIC) displacement field measurements. As shown in Figure \ref{fig:dicsetup}, a material database is constructed from the DIC displacement tracking measurements of seven biaxial stretching protocols with different biaxial tension ratios on a porcine tricuspid valve anterior leaflet specimen. Then, the material response is modeled as a solution operator from the loading to the resultant displacement field, using the implicit FNO \cite{you2022learning} architecture, to predict the response of this soft tissue specimen with arbitrary loading conditions.

To demonstrate the predictivity, we use various combinations of loading protocols, and compare its effectiveness with a finite element analysis approach based on the phenomenological Fung-type model. To study the model performance for in-distribution tests, we randomly selected $83\%$ of the samples of all protocols to form the training set, and built the vanilla IFNO model and the Fung-type model based on this common training set. Then, both models are validated and tested on the rest of the samples. Moreover, to further investigate the generalization capability of the models, we also perform the out-of-distribution prediction study, but employing part of the protocols for training and the rest of protocols for testing. In particular, the models are trained on protocols with biaxial tensions $P_{11}:P_{22}=1:1$, $P_{11}:P_{22}=0.66:1$, and $P_{11}:P_{22}=1:0.66$, then tested on protocols with $P_{11}:P_{22}=1:0.33$, $P_{11}:P_{22}=0.33:1$, $P_{11}:P_{22}=0.1:1$, and $P_{11}:P_{22}=1:0.05$. We notice that the testing protocols are not covered in any of training sets, and they have smaller maximum tensions compared with the training sets. Hence, with this study we aimed to investigate the performance of the implicit FNO and the physics-guided implicit FNO methods for predicting the out-of-distribution material responses in the small deformation regime.

\paragraph{Results: } From in-distribution tests (see Figure \ref{fig:unsmoothdata_study1}(a) and (b)), we found that the proposed data-driven approach presents good prediction capability to unseen loading conditions with the same type of biaxial loading ratios and outperforms the phenomenological model. Specifically, The implicit FNO model achieved only $1.64\%$ relative error on the test dataset, while the Fung-type model has a $10.83\%$ error. To provide further insights into this comparison, in Figure \ref{fig:unsmoothdata_study1}(b) both the {\it x}- and {\it y}-displacement solutions and the prediction errors are visualized on a representative test sample. The Fung-type model, which considered the homogenized stress--strain at one material point (i.e., the center of the specimen) due to limited information about the spatial variation in the stress measurement,  failed to capture the material heterogeneity and hence exhibited large prediction errors in the interior region of the TVAL specimen domain. This observation confirms the importance of capturing the material heterogeneity and verifies the capability of the neural operators in heterogeneous material modeling for in-distribution learning tasks.

On the other hand, when tested on out-of-distribution loading ratios, the neural operator learning approach becomes less effective and has a comparable performance as the constitutive modeling (see Figure \ref{fig:unsmoothdata_study1}(c) and (d)). 
In particular, when the models are trained on protocols in the large deformation region then tested on protocols in the small deformation region, $16.78\%$ and $16.80\%$ prediction errors were observed from the implicit FNO and Fung-type model, respectively. To improve the generalizability of the neural operators, partial physics knowledge was infused using the no-permenant-set assumption discussed in Section \ref{sec:pifno}, and this method is shown to improve the model's extrapolative performance in the small deformation regime by around $1.5\%$. This study demonstrates that with sufficient data coverage and/or regularization terms from partial physics constraints, the data-driven neural operator learning approach can be a more effective method for modeling complex biological materials than the traditional constitutive modeling approaches. The results for this problem have been adapted from \cite{you2022physics}.

\section{Summary and Outlook}
\label{sec:summary}

In this chapter, we have reviewed the basics of three neural operators, DeepONet, FNO and the graph neural operator, as well as their extensions, and have presented representative application examples. While the list of possible applications of neural operators will continue to expand in the near future, here we provide a partial list of their role in applications so far.

\begin{itemize}
    \item For real time forecasting: designing efficient control systems, fault detectors for car engines, solving complex multi-physics problems in less than a second \cite{cai2021deepm}.
    \item Proper hybridization of physics- and data-based models can achieves the goal of generating an efficient, accurate, and generalizable model that can be used to greatly accelerate modeling of time-dependent multi-scale systems \cite{yin2022interfacing}. 
    \item To reduce the dependency of large paired datasets (when accurate information about the governing law of the physical system is not available), Deep Transfer Operator Learning (DeepONet)\cite{goswami2022deep} can be used for accurate prediction of quantities of interest in related domains, with only a handful of new measurements.  
    \item Develop faster ways to train neural operators by incorporating  multimodality and multifidelity data \cite{lu2022multifidelity,howard2022multifidelity,de2022bi}.
    \item The application of neural operators in life sciences is endless. For example, approaches developed in \cite{yin2022simulating,goswami2022neural} show the application of DeepONet for accurate prediction of aortic dissection and aneurysm, which is patient-specific and hence could provide clinicians sufficient time for planning a surgery.
    \item DeepONet can be used for accelerating climate modeling by adding learned high-order corrections to the low resolution (e.g., 100 Km) climate simulations. 
\end{itemize}

\begin{figure}[!htb]
    \centering 
    \subfigure[]{\includegraphics[scale=0.4]{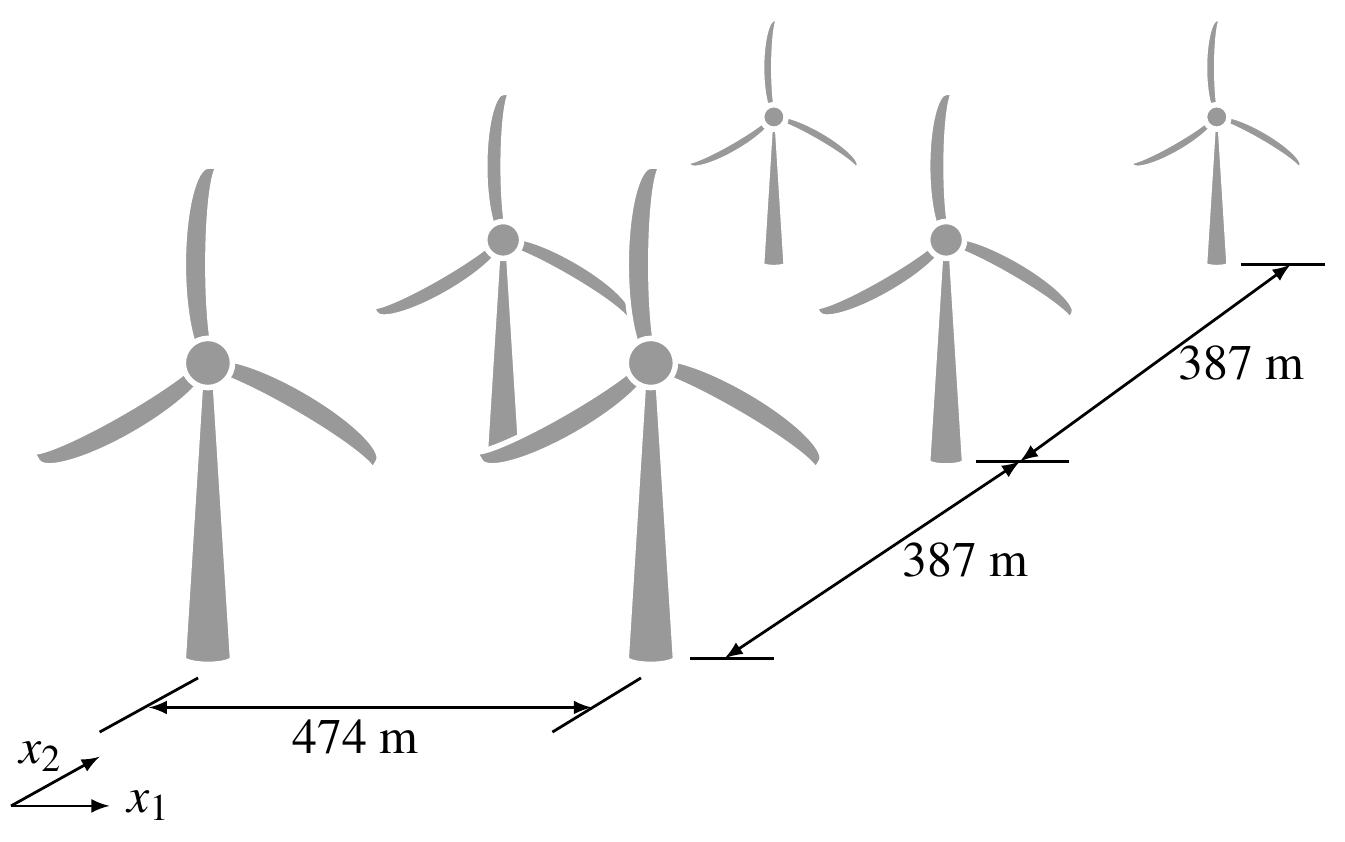}}
    \subfigure[]{\includegraphics[scale=0.18]{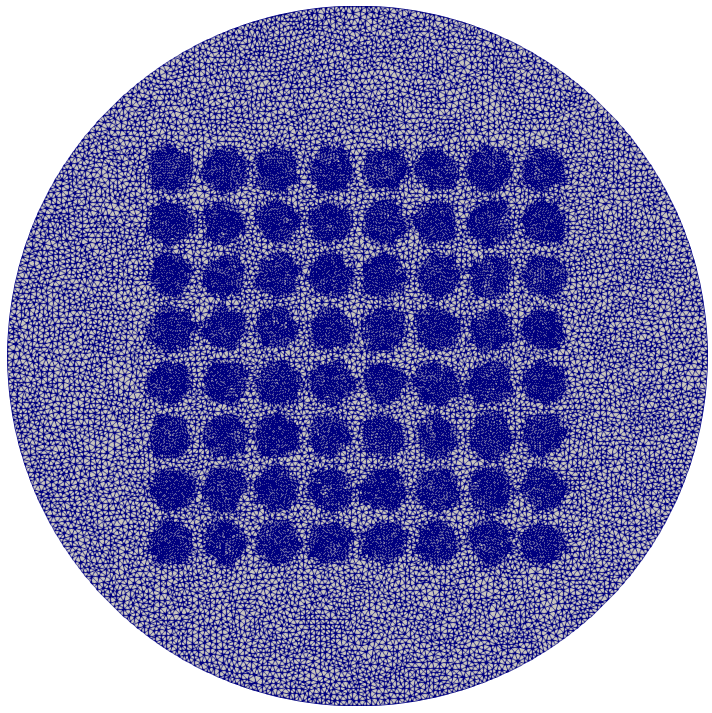}}
    \caption{(a)Wind farm layout of 6 wind turbines. (b) Arrangements of $8\times8$ wind turbines in a farm. Adopted from \cite{de2022bi}.} 
    \label{fig:wind_farm}
\end{figure}

Next, we discuss possible new developments required in the future to further advance physics-informed deep learning, and in particular neural operators.

In the last two decades, we have seen rapid advances in GPU computing that together with the simultaneous advances in deep learning algorithms have enabled the development of new hybrid models based on both physics-driven and data-driven methods. Research teams are now working on developing high-fidelity digital twins of the human organs and the Earth's atmosphere, which will require long and expensive training, even more than the expensive transformer language models developed by the big software companies involving hundreds of billions of parameters.  To deal with the increasing cost and cope with the urgent demand for real-time inference that requires only very little new training for problems at scale in computational mechanics,  higher levels of abstraction of these algorithms is required. To that end, continual learning at the operator regression level is a promising avenue  in establishing the mathematical foundations of digital twins in computational mechanics and beyond. Some authors and even industry researchers use concepts from principal component analysis and reduced order modeling to build digital twins but  the lack of overparametrization of such methods, even of their nonlinear extensions, is a limiting factor for their effectiveness in realistic scenarios of diverse and unanticipated operating conditions. 

Adding physics into the training of neural operators in addition to any available data enhances their accuracy and generalization capacity for tasks even outside the distribution of the input space.   Scalable physics-informed neural networks \cite{shukla2022scalable} can be employed to solve high-dimensional problems not possible with traditional finite element solvers, e.g., up to approximately 10 dimensions if not more. Similarly, scalable physics-informed neural operators can also solve high-dimensional problems even in real time and can be used for designing very complex systems. For example,  in \cite{de2022bi} the authors solved an industry-based problem of computing the power generated in a utility-scale wind plant with $64$ turbines, considering the uncertainty of the wind speed, inflow direction, and yaw angle (a schematic representation shown in Figure \ref{fig:wind_farm}). The layout of $64$ wind turbines is on a two-dimensional mesh which was used to generate the training data as shown in Figure~\ref{fig:wind_farm}(b). The annual energy output of a wind farm is often calculated by estimating the predicted power with regard to the joint distribution of wind speed and direction; however, the quantity of function evaluations necessary frequently prevents the use of high-fidelity models in industry \cite{king2020probabilistic}. 

To that end, training a DeepONet or any other neural operator with only high fidelity data would be computationally expensive. One promising way to solve such realistic problems is using multifidelity approaches proposed in \cite{de2022bi,lu2022multifidelity,howard2022multifidelity}. These real problems take leverage of the generalized flavor of DeepONet,  which can be flexibly designed for any problem at hand. Scaling of neural operators to industry-level problems with parallel multi-GPU training could be a very impactful research direction. Another interesting direction is developing  graph neural operators for modeling digital twins and complex systems-of-systems, in particular, with the ability to apply causal inference for discovering intrinsic pathways not easily discovered by other methods. Finally, to address the excessive cost of training in deep learning and make edge computing a reality, developing the energetically favorable spiking neural networks on neuromorphic computers could lead to (at least) three orders of magnitude in energy savings while we may come closer to more biologically plausible neural operator architectures; see also Figure 1. 

\bibliographystyle{elsarticle-num} 
\bibliography{biblio}

\end{document}